\documentclass{article}
\usepackage[utf8]{inputenc}
\usepackage{microtype}
\usepackage{graphicx}
\usepackage{subfigure}
\usepackage{booktabs} 
\usepackage{comment}
\usepackage{colortbl}
\usepackage{arydshln}
\usepackage{makecell}
\usepackage{hyperref}
\usepackage{mathtools}
\usepackage{color}
\usepackage{graphicx}
\usepackage{CJKutf8}
\usepackage{url}
\usepackage{multicol,multirow}
\usepackage{makecell}
\usepackage{array}
\usepackage{bm}
\usepackage{parskip}
\usepackage[T1]{fontenc}    
\usepackage{booktabs}       
\usepackage{amsmath, amsthm}
\usepackage{amssymb}
\usepackage{amsfonts}
\usepackage{times}
\usepackage{latexsym}
\usepackage{amsmath, amsthm}
\usepackage{amssymb}
\usepackage{amsfonts}

\usepackage{geometry}

\definecolor{ao}{rgb}{0.0, 0.5, 0.0}
\definecolor{asparagus}{rgb}{0.53, 0.66, 0.42}
\definecolor{amber}{rgb}{1.0, 0.49, 0.0}
\definecolor{alizarin}{rgb}{0.82, 0.1, 0.26}
\definecolor{applegreen}{rgb}{0.55, 0.71, 0.0}
\definecolor{amethyst}{rgb}{0.6, 0.4, 0.8}
\definecolor{auburn}{rgb}{0.43, 0.21, 0.1}

\title{Interpreting Deep Learning Models in Natural Language Processing: \\ A Review}
\author{Xiaofei Sun$^{1}$, 
Diyi Yang$^{2}$, 
Xiaoya Li$^{1}$,
Tianwei Zhang$^{3}$, \\
Yuxian Meng$^{1}$, 
Han Qiu$^{4}$,
Guoyin Wang$^{5}$, 
Eduard Hovy$^{6}$,
Jiwei Li$^{1,7}$ \\\\
$^{1}$Shannon.AI, $^{2}$Georgia Institute of Technology\\
$^{3}$Nanyang Technological University, $^{4}$Tsinghua University \\
$^{5}$Amazon Alexa AI, $^{6}$Carnegie Mellon University, $^{7}$Zhejiang University \\
}

\date{}

\begin{document}

\maketitle

\begin{abstract}
Neural network models have achieved state-of-the-art performances in a wide range of natural language processing (NLP) tasks. However, a long-standing criticism against neural network models is the lack of interpretability, which not only reduces the reliability of neural NLP systems but also limits the scope of their applications in areas where interpretability is essential (e.g., health care applications). In response, the increasing interest in  interpreting neural NLP models has spurred a diverse array of interpretation methods  over recent years. In this survey, we provide a comprehensive review of various interpretation methods for neural models in NLP. We first stretch out a high-level taxonomy for interpretation methods in NLP, i.e., training-based approaches, test-based approaches and hybrid approaches. Next, we describe sub-categories in each category in detail, e.g., influence-function based methods, KNN-based methods, attention-based models, saliency-based methods, perturbation-based methods, etc. We point out  deficiencies of current methods and suggest some avenues for future research. 
\footnote{E-mail addresses: \url{xiaofei_sun@shannonai.com} (X. Sun), \url{diyi.yang@cc.gatech.edu} (D. Yang), 
\url{xiaoya_li@shannonai.com} (X. Li), \url{tianwei.zhang@ntu.edu.sg} (T. Zhang), \url{yuxian_meng@shannonai.com} (Y. Meng), \url{qiuhan@tsinghua.edu.cn} (H. Qiu), \url{guoyiwan@amazon.com} (G. Wang), \url{hovy@cmu.edu} (E. Hovy), \url{jiwei_li@shannonai.com} (J. Li).}
\end{abstract}

\section{Introduction}
\label{introduction}
Deep learning based models have achieved state-of-the-art results on a variety of natural language processing (NLP) tasks such as 
tagging \cite{ma2016end,shao2017character,lattice2018zhang,li2019unified}, 
text classification \cite{kim2014convolutional,zhou2015c,radford2018improving,lin2021bertgcn,chai2020description, wang2018joint}, 
machine translation \cite{vaswani2017transformer,edunov2018understanding,zhu2020incorporating,takase2021lessons}, natural language understanding \cite{yang2019xlnet,lan2019albert,sun2019ernie,wang2021entailment},
dialog  \cite{adiwardana2020towards,xu2018dp,baheti2018generating}, 
 and question answering \cite{danqi2020spanbert,karpukhin2020dense,zhang2020retrospective,yamada2020luke}. By first transforming each word (or each {\it token} in the general sense) into its word vector and then mapping these vectors into higher-level representations through layer-wise interactions such as
 recursive nets \cite{socher2013recursive,irsoy2014deep,li2015tree,bowman2016fast},  LSTMs  \cite{hochreiter1997long}, CNNs \cite{kalchbrenner2014convolutional,kim2014convolutional}, Transformers \cite{vaswani2017transformer},
  deep neural networks are able to encode rich linguistic and semantic knowledge in the latent vector space \cite{mikolov2013distributed,clark2019does,htut2019attention,hao-etal-2019-visualizing}, capture contextual patterns and make accurate predictions on downstream tasks.
However, 
a long-standing criticism against deep learning models is the lack of interpretability: 
in NLP, it is unclear how neural models 
 deal with {\it composition} in language, such as implementing affirmation, negation, disambiguation and semantic combination from different constituents of the sentence \cite{jain2019attention,wiegreffe-pinter-2019-attention,talmor2019olmpics}. The {\it uninterpretability} of deep neural NLP models limits their scope of applications 
 which require strong 
 controllability and interpretability. 
 
Designing methods to interpret neural networks has gained increasing attentions. 
in the field of computer vision, 
interpreting neural models involves identifying the part of the input that contributes most to the model predictions, and this paradigm has been widely studied  (CV),
using  saliency maps \cite{simonyan2013deep,yosinski2015understanding}, layer-wise relevance propagation \cite{bach2015lrp,montavon2017explaining,kindermans2017learning}, model-agnostic interpreting methods \cite{ribeiro2016i} and back-propagation \cite{shrikumar2017learning}. These methods have improved neural model interpretability and reliability from various perspectives, and  raised the community's awareness of the necessity of neural model interpretability when we want to build trustworthy and controllable systems \cite{adadi2018peeking,tjoa2020survey,miller2019explanation}.
Different from computer vision, 
the basic input units in NLP for neural  models are discrete language tokens rather than continuous pixels in images
\cite{10.5555/944919.944966,cho-etal-2014-learning,bahdanau2014neural}. This discrete nature of  language poses a challenge for interpreting neural NLP models, making the interpreting methods in CV hard to be directly applied to NLP domain \cite{bodria2021benchmarking,lertvittayakumjorn2021explanationbased}.
 To accommodate the discrete nature of texts,
  a great variety of  works have rapidly emerged over the past a few years for neural model interpretability. 
 A systematic review is thus needed to sort out and compare different methods, providing a comprehensive understanding of from which perspective a neural NLP model can be interpreted and how the model can be interpreted effectively.
\begin{table}[t]
    \centering
    \begin{tabular}{lp{10cm}}
        \toprule 
        {\bf Training-based methods} & Identify  training instances 
        responsible for the prediction of the current test instance.\\
        {\bf Test-based methods} & Identify which part(s) of a test instance responsible for the model prediction.\\
        {\bf Hybrid methods} & Identify because of which training instances, the model attends to which part(s) of the test instance for its prediction.\\
        \bottomrule
    \end{tabular}
    \caption{Summary of the high-level categories of interpretation methods for neural NLP models.}
    \label{tab:summary}
\end{table}

\begin{figure}[t]
    \centering
    \includegraphics[width=1\textwidth]{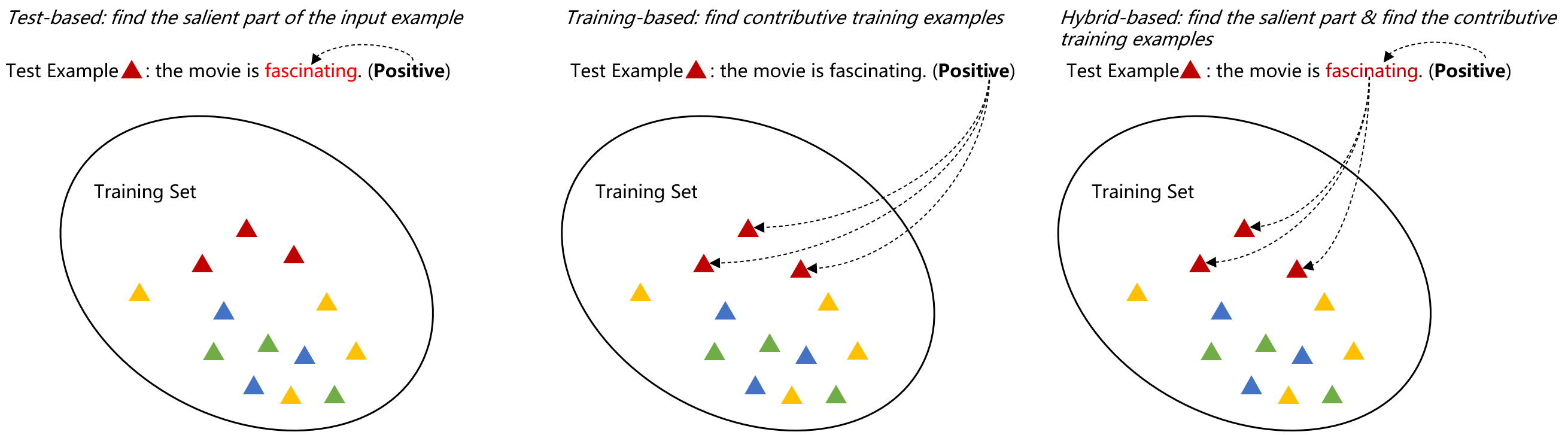}
    \caption{Difference of training-based, test-based and hybrid interpreting methods. The figure is adapted from the Figure 1 in the work of \cite{meng2020pair}. In a nutshell, test-based methods target finding the most salient part(s) of the input; training-based methods find the most contributive training examples responsible for a model's predictions; hybrid methods combine both worlds, by first finding the most salient part(s) of the input and then identifying the training instances responsible for the salient part(s).}
    \label{fig:overview}
\end{figure}

\begin{figure}[t]
    \centering
    \includegraphics[width=0.6\textwidth]{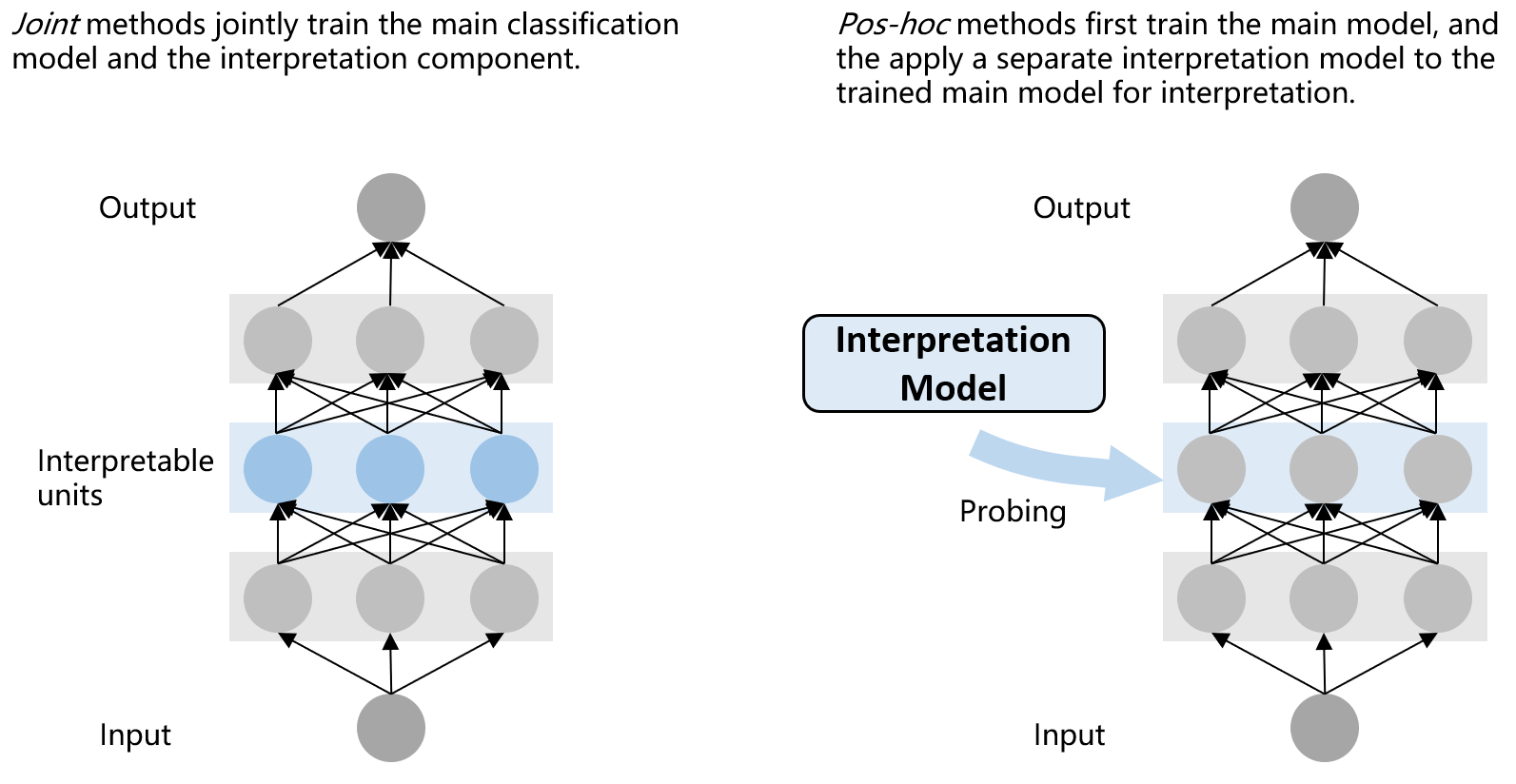}
    \caption{An illustrative comparison between {\it joint} interpreting methods and {\it post-hoc} interpreting methods. }
    \label{fig:joint-post}
\end{figure}

In this survey, we aim to to provide a comprehensive review of existing interpreting methods for neural NLP models. We 
propose  taxonomies 
to categorize 
 existing interpretation methods 
 from  the perspective of (1)   {\it training-focused} v.s.   {\it test-focused}; 
 and (2)   {\it joint}  v.s. {\it post-hoc}. 
Regarding training-focused v.s. test-focused, 
the former denotes identifying the training instances 
or elements within training instances responsible for the model's  prediction on a specific test example, while the latter denotes  
 finding
the most salient part of the input test-example responsible for a prediction. 
The two perspectives
 can be jointly considered: aiming to identify because of which training instances, the model attends to which part(s) of the test example for its prediction. 
 Based on the specific types of algorithms used, 
  training-based methods can be further divided into influence-function based methods, KNN-based methods,  kernel-based methods, etc. 
 Test-based methods can be  divided into saliency-based methods,
 perturbation-based methods, attention-based methods, etc. 
 Figure \ref{fig:overview} depicts the difference of training-based, test-based and hybrid interpreting methods.
Regarding  joint  v.s.  post-hoc, 
for the former joint appraoch, 
a  joint interpreting model is  trained along with the main classification model and the interpreting tool is nested within the main model, such as attention;
for the latter post-hoc approach, 
 an additional probing model, which is separately developed  
  after the main model completes training, is used for interpretation. 
   Figure \ref{fig:joint-post} depicts the difference between joint and post-hoc approaches. 
 Categorization with taxonomy framework naturally separates different methods and makes audience easy to locate the works of interest. 
Table \ref{tab:summary} shows the high-level categories of explanations

We analyze the literature with respect to these dimensions and review some of the  representative techniques regarding each of these categories. 
In summary, this work:
\begin{itemize}
    \item Categorizes interpreting methods according to {\it training-focused} v.s.   {\it test-focused}
 and   {\it joint}  v.s. {\it post-hoc}.
    \item Provides an empirical review of up-to-date  methods for interpreting neural NLP models.
    \item Discusses deficiencies of current methods and suggest  avenues for future research.
\end{itemize}

The rest of this survey is organized as follows: Section \ref{taxonomy} clarifies the high-level taxonomy proposed in this work. Section \ref{training}, \ref{test} and \ref{hybrid} review existing training-based methods, test-based methods and hybrid methods for interpreting neural NLP models. Section \ref{direction} discusses some open problems and directions that may be intriguing for future works. Section \ref{conclusion} summarizes this survey.

\subsection{Related Surveys}

Related to this work, \cite{zhang2021survey}  conducted a survey of neural network interpretability methods used in computer vision, bioinformatics and NLP along several different dimensions such as using passive/active interpretation approaches or focusing on local/global effect.  However, our survey provides an in-depth overview for interpreting neural NLP models for NLP tasks. 
Another line of work, AllenNLP Interpret \cite{wallace2019allennlp},  provides interpretation primitives (e.g., input gradients) for AllenNLP model and task, a suite of built-in interpretation methods, and a library of front-end visualization components;
the toolkit mainly supports  gradient-based saliency maps and adversarial attacks. In contrast, our review covers more comprehensive interpretability methods. 

\section{High-level Taxonomy}
\label{taxonomy}
\begin{table}[t]
    \centering
    \small
    \begin{tabular}{llp{6cm}p{1.7cm}p{3cm}}
        \toprule 
        {\bf First Level} & {\bf Second Level} & {\bf Technique Description} & {\bf Joint/Post-hoc} & {\bf Representative Works}\\\midrule
        & Influence functions & Measuring the change in model parameters when a training point changes. & Post-hoc & \cite{han2020explaining,kobayashi-etal-2020-efficient,yang2020generative,guo2020fastif}\\
        \multirow{3}{*}{\bf Training-based} & KNNs for interpretation & Searching $k$ nearest neighbors over a model's hidden representations to identify training examples closest to a given evaluation example.  & Post-hoc &\cite{wallace2018interpreting,rajani2020explaining}\\
        & Kernel-based explanation & Within the kernel-based framework, selecting landmarks that the Layer-wise Relevance Propagation algorithm highlights as the most active elements responsible for the model prediction. & Joint/Post-hoc & \cite{croce-etal-2018-explaining,croce2019auditing}\\\hline
        & Saliency maps & Computing the saliency score of each token or span based on network gradient or feature attribution. The part of sentence with the highest salience score is viewed as the interpretable content. & Post-hoc & \cite{li2015visualizing,li2016understanding,sundararajan2017axiomatic,ross2017right,ancona2017towards,poerner-etal-2018-evaluating,feng2018pathologies,guan2019towards,kim-etal-2020-interpretation,chen-ji-2020-learning}\\
        \multirow{3}{*}{\bf Test-based} & Attention as explanation & Leveraging attention distributions over tokens as a tool of interpretation. The most attentive token or span is regarded as the most contributive to the model prediction. & Joint & \cite{sun2020selfexplaining,jain2019attention,serrano2019attention,wiegreffe-pinter-2019-attention,vashishth2019attention,rogers2020primer,ghaeini2018interpreting,clark2019does,vig2019analyzing,tenney2019bert,htut2019attention,Brunner2020On,pruthi2019learning,hao-etal-2019-visualizing,reif2019visualizing}\\
        & Explanation generation & Generating or inducing explanations for the model's prediction. The explanation may come from a subset of the input sentence, from external knowledge or from a language model that generates explanations from scratch. & Joint & \cite{lei2016rationalizing,DBLP:journals/corr/abs-2004-14546,chang2019game,swanson2020rationalizing,deyoung-etal-2020-eraser,jain-etal-2020-learning,kumar2020nile,rajani2019explain,liu2018towards,dong-etal-2019-editnts,chrysostomou2021variable,Huang_Chen_Du_Yang_2021,bastings2019interpretable,schlichtkrull2020interpreting,de2020decisions,ribeiro2018anchors,rajagopal2021selfexplain}\\\hline
        \multirow{1}{*}{\bf Hybrid} & & Interpreting model behaviors by jointly examining training history and test stimuli. & Post-hoc & \cite{meng2020pair}\\
        \bottomrule
    \end{tabular}
    \caption{An overview of high-level taxonomy for interpretation methods in NLP. Column 1 denotes the topmost level of the taxonomy. Column 2 denotes the second-level categorizations. Column 3 gives brief descriptions of the techniques grouped in the second-level categorizations. Column 4 notifies whether the body of work is {\it joint} (the interpreting model is trained jointly with the main model) or {\it post-hoc} (the interpreting method is applied to the trained main model). Column 5 contains a list of representative references.}
    \label{tab:taxonomy}
\end{table}

In this section, we provide a high-level overview of the taxonomy used in this survey. As shown in Table \ref{tab:taxonomy}, the topmost level categorizes existing methods by whether they aim to interpret models from the training side, from the test side, or from both sides.

Walking into the second level, this taxonomy categorizes literature based on the basic tools they use to interpret model behaviors. For training-based methods, they are sub-categorized into influence functions (Section \ref{influence}), KNNs for interpretation (Section \ref{knn}) and kernel-based explanation (Section \ref{kernel}), which are respectively related to using influence functions, $k$ nearest neighbors and kernel-based models for interpretation in the training side. Test-based methods (Section \ref{test}) are sub-categorized into saliency maps (Section \ref{saliency}), attention as explanation (Section \ref{attention}), and explanation generation (Section \ref{generation}), which respectively use saliency maps, attention distributions and explainable contexts to interpret models in the test side. Hybrid methods interpret model predictions by jointly examining training history and test stimuli, which can be viewed as a combination of training-based and test-based methods (Section \ref{hybrid}). 
Each category is paired with whether they belong to  {\it joint}  or {\it post-hoc}. 
In the following section of this survey, we will expound on some of the representative works from each of the categories.

\section{Training-Based Interpretation}
\label{training}
The main purpose of training-based methods is to identify  which training instances or units of training instances 
in the training set 
are  responsible for the model's prediction at a specific test example.  Existing 
training-based
techniques fall into the following three categories:
\begin{itemize}
  \item {\bf Influence function based interpretation}: 
  using influence functions to estimate the {\it influence} of each training point on an input test example. The training points with the highest influences are viewed as the most contributive instances. 
    \item {\bf Nearest Neighbor based interpretation}:  
  extracting the $k$ nearest neighbor  points in the training set that
  are close to an input test example.
Extracted  nearest neighbors  are viewed as the most contributive instances
    \item {\bf Landmark based interpretation}:
    An input test sentence is fed to the classifier along with {\it landmarks}, which refer to a set of training examples used to compile the representation of any unseen test instance.
    The influence of training examples on an input test example is measured by
    the contribution of each landmark to the model's decision on the test example based on the Layer-wise Relevance Propagation (LRP) \cite{layer-relev-propagation} algorithm. 
As landmarks are organized via kernel-based deep architectures, this method is also referred to as {\it Kernel-based Explanation}.
\end{itemize}
We will describe these methods at length in the rest of this subsection.

\subsection{Influence Functions Based Interpretation}
\label{influence}
\subsubsection{Influence functions}
The {\it influence} of a training point can be defined as the change in the model parameters when 
a specific training point is removed.
To measure the change of model parameters,  
an intuitive way is to retrain the model on the training set with the training example left-out. 
However, this is prohibitively time-intensive as we need to retrain the model from scratch for every training example. 
Alternatively, {\it influence functions} \cite{cook1982residuals,cook1986assessment,koh2017understanding} provide an approximate but tractable way 
to track parameter change 
 without the need to repeatedly retrain the model.

Concisely speaking, the influence function estimates the impact on the model parameters when the loss of a particular training point is up-weighted by a small perturbation $\epsilon$. This can be done through first-order approximation. 
Let $z_i (i=1,2,\cdots,n)$ be a training point in the training set. The model parameters $\theta$ is learned to minimize a given loss function $L$:
\begin{equation}
  \hat{\theta}=\arg\min_\theta\frac{1}{n}\sum_{i=1}^nL(z_i;\theta)
  \label{eq:minimize}
\end{equation}
Analogously, the optimal model parameters $\theta_{z,\epsilon}$ when the loss of a training point $z$ is weighted by a small perturbation can be derived by minimizing the following loss:
\begin{equation}
  \hat{\theta}_{z,\epsilon}=\arg\min_\theta\frac{1}{n}\sum_{i=1}^nL(z_i;\theta)+\epsilon L(z;\theta)
  \label{eq:perturbation}
\end{equation}
By referring the audience to \cite{koh2017understanding} for detailed proofs, the influence of a training point $z$ on the model parameters $\theta$, which is denoted by $I(z,\theta)$, is given as follows:
\begin{equation}
  I(z,\theta)\triangleq\frac{\mathrm{d}\hat{\theta}_{z,\epsilon}}{\mathrm{d}\epsilon}=-H_\theta^{-1}\nabla_\theta L(z;\hat{\theta})
  \label{eq:influence}
\end{equation}
where $H_\theta\triangleq\frac{1}{n}\sum_{z_i}\nabla_\theta^2L(z_i;\hat{\theta})$ is the Hessian matrix over all training instances $z_i$ in the training set, and $L(z_i;\hat{\theta})$ is the loss of $z$ over model parameters $\theta$. 
Equation \ref{eq:influence} expresses the degree to which a small perturbation on a training point $z$ influences the model parameters $\theta$. Based on Equation \ref{eq:influence}, \cite{han2020explaining} applied the chain rule to measure how this change in model parameters would in turn affect the loss of the test input $x$:
\begin{equation}
  I(z,x)\triangleq\frac{\mathrm{d}L(x;\hat{\theta})}{\mathrm{d}\epsilon_z}=\nabla_\theta L(x;\hat{\theta})\cdot \frac{\mathrm{d}\hat{\theta}_{z,\epsilon}}{\mathrm{d}\epsilon}=\nabla_\theta L(x;\hat{\theta})\cdot I(z,\theta)=-\nabla_\theta  L(x;\hat{\theta})^\top H_\theta^{-1}\nabla_\theta L(z;\hat{\theta})
  \label{eq:influence-chain}
\end{equation}
\cite{han2020explaining} defined the negative score $-I(z,x)$ as the {\it influence score}.
Equation \ref{eq:influence-chain} reveals that a positive influence score for a training point $z$ with respect to the input example $x$ means that the removal of $z$ from the training set would be expected to cause a drop in the model's confidence when making the prediction on $x$. By contrast, a negative influence score means that the removal of $z$ from the training set would be expected to give an increase in the model's confidence when making this prediction.
Experiments and analysis from \cite{han2020explaining} validated the reliability of influence functions when applied to deep transformer-based \cite{vaswani2017transformer} models.

\subsubsection{Efficient influence functions}
\begin{figure}[t]
  \centering
  \begin{minipage}[t]{0.48\textwidth}
    \centering
    \includegraphics[width=1\textwidth]{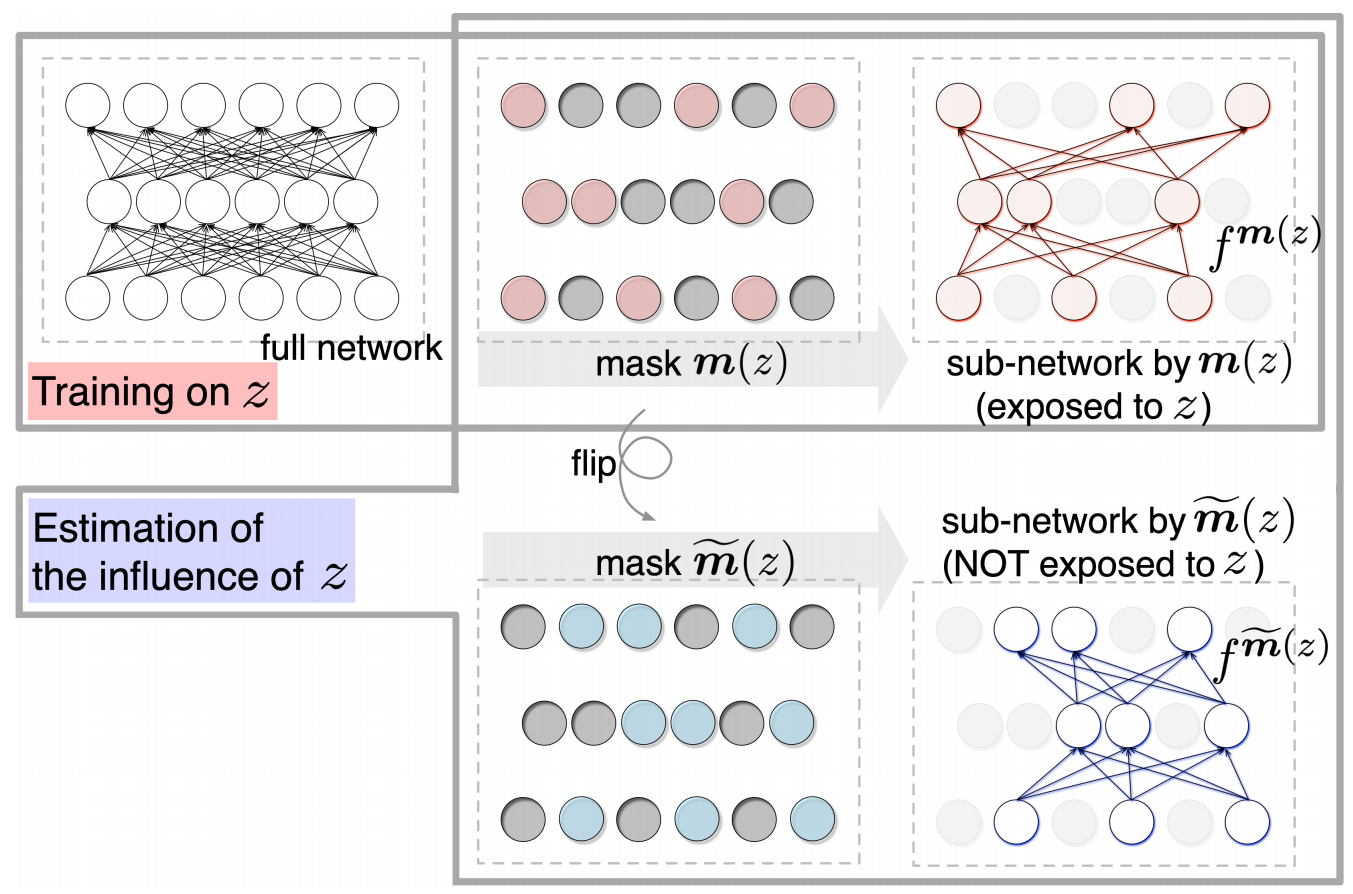}
  \end{minipage}%
  \hfill
  \begin{minipage}[t]{0.48\textwidth}
    \centering
    \includegraphics[width=1\textwidth]{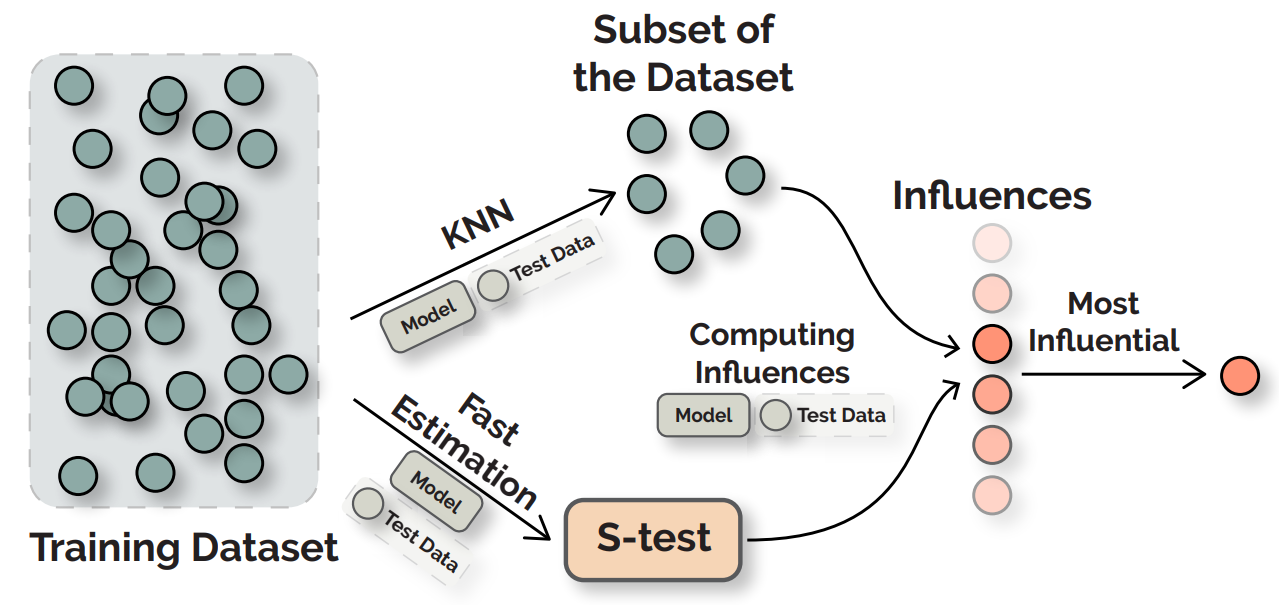}
  \end{minipage}
  \caption{Efficient implementations of influence functions proposed in \cite{kobayashi-etal-2020-efficient} (left) and \cite{guo2020fastif} (right). The figures are borrowed from the original literature.}
  \label{fig:influence}
\end{figure}

A closer look at Equation \ref{eq:influence-chain} reviews the disadvantage of influence functions: the computational cost of computing the inverse-Hessian matrix $H_\theta^{-1}$ and iterating over all training points $z$ to find the most (least) influential example is infeasible to bear for large-scale models such as BERT \cite{devlin2018bert}.  Efficient estimates for influence functions are thus needed to improve runtime efficiency while preserving the interpretability.
\cite{kobayashi-etal-2020-efficient} and \cite{guo2020fastif} respectively address this issue from the model side and from the algorithm side. Figure \ref{fig:influence} illustrates how these two methods work.
\cite{kobayashi-etal-2020-efficient} proposed {\it turn-over dropout}, a variant of dropout  applied to the network for each individual training instance. After training the model on the entire training dataset, {\it turn-over dropout} generates an individual-specific dropout mask scheme $\bm{m}(z)$ for each training point $z$, and then feeds $z$ into the model for training using the proposed mask scheme $\bm{m}(z)$. The mask scheme $\bm{m}(z)$ corresponds to a sub-network $f^{\bm{m}(z)}$, which is updated when the model is trained on $z$, but the counter-part of the model $f^{\tilde{\bm{m}}(z)}$, is not at all affected. Therefore, the two sub-networks,  $f^{\bm{m}(z)}$ and $f^{\tilde{\bm{m}}(z)}$, can be analogously perceived as two different networks trained on a dataset with or without $z$. The influence of $z$ on a test input example $x$ can thus be computed as follows:
\begin{equation}
  I(z,x)\triangleq L(x;f^{\bm{m}(z)})-L(x;f^{\tilde{\bm{m}}(z)})
  \label{eq:influence-dropout}
\end{equation}
Experimenting on the Stanford Sentiment TreeBank (SST-2)  \cite{socher-etal-2013-recursive} and the Yahoo Answers dataset \cite{zhang2015character} with the BERT model, the {\it turn-over dropout} technique is able to interpret erroneously classified test examples by identifying the most influential training instances, with a more efficient computation process.
\cite{guo2020fastif} tackled the computation issue of the original influence functions from the algorithm perspective. Their optimization process can be divided into three aspects: (1) $k$NN constraint on the search space; (2) the inverse Hessian computation speedup; and (3) parallelization. First, to avoid the global search on the entire training dataset for the most influential data point, the search space is constrained within a small subset of the training set. This can be achieved by selecting the top-$k$ nearest neighbors to the test input instance $x$ based on the $l_2$ distance between extracted features. Second, to speed up computation of the inverse Hessian matrix $H_\theta^{-1}$, \cite{guo2020fastif} carefully selected the required hyperparameters used to compute the inverse Hessian matrix so that the inverse Hessian-vector product $H_\theta^{-1}\nabla_\theta L(x;\hat{\theta})$ (Equation \ref{eq:influence-chain}) can efficiently calculated. Third, \cite{guo2020fastif} applied off-the-shelf multi-GPU parallelization tools to compute influence scores in a parallel manner. The confluence of all three factors leads to a speedup rate of 80x while being highly correlated with the original influence functions.
With the new fast version of influence functions, a few applications that are concerned with model interpretation but were previously intractable are demonstrated, including examining the ``explainability'' of influential data-points, visualizing data influence-interactions, correcting model predictions using original training data, and correcting model predictions using data from a new dataset. The efficient implementation of influence functions provide possibility for new interpretable explorations that scale with the data size and model size.

The influence function based interpreting methods are along the {\it post-hoc} line, which means that they are applied for interpretation after the main model is fully trained.

\subsection{KNNs Based Interpretation}
\label{knn}
\begin{figure}[t]
    \centering
    \begin{minipage}[t]{0.48\textwidth}
      \centering
      \includegraphics[width=1\textwidth]{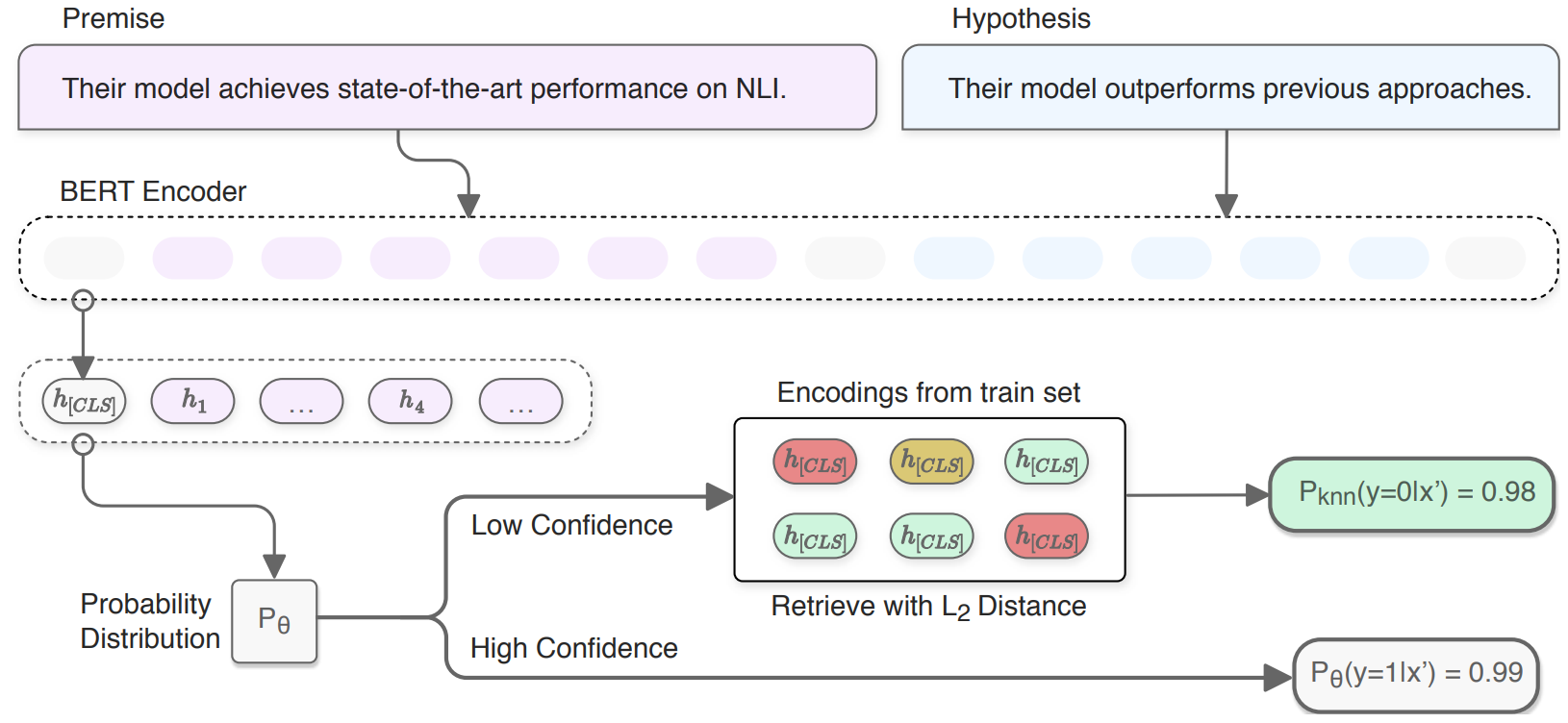}
    \end{minipage}%
    \hfill
    \begin{minipage}[t]{0.48\textwidth}
      \centering
      \includegraphics[width=1\textwidth]{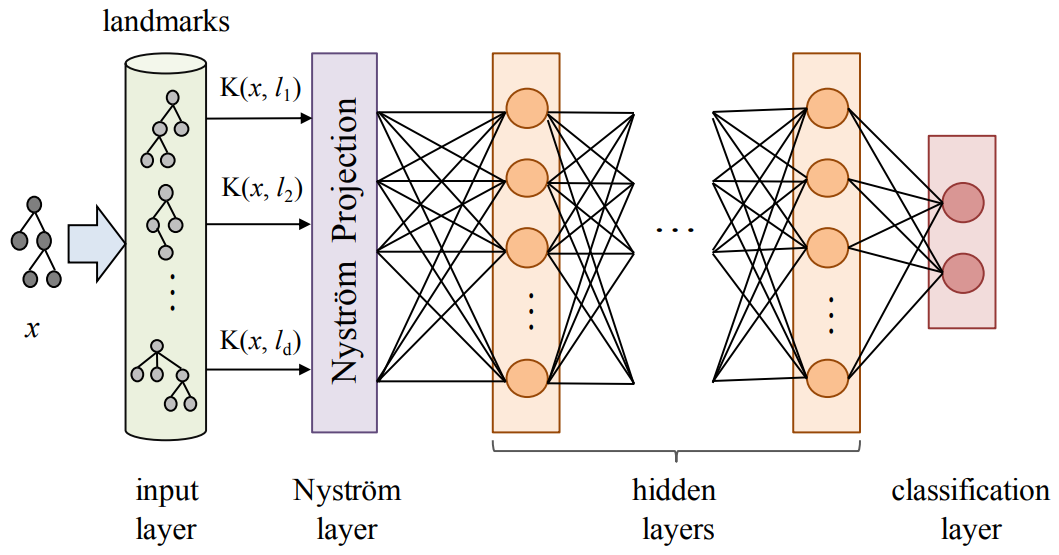}
    \end{minipage}
    \caption{Overview of KNN-based (left, the figure is borrowed from the work of \cite{rajani2020explaining}) and kernel-based (right, the figure is borrowed from the work of \cite{croce2019auditing}) interpretation methods. KNN-based methods first cache high-dimensional representation of training instances and search for the $k$ nearest neighbors with respect to the test example in the latent space. Kernel-based methods leverage {\it landmark} instances from the training set as part of the model input, providing linguistic evidence for interpretation.}
    \label{fig:knn_kda}
\end{figure}

KNN-based interpretation \cite{wallace2018interpreting,rajani2020explaining} retrieves 
$k$ nearest training instances that are the closest to a specific test input 
from the training set. After the training process completes, each training point is passed to the trained model to derive its high-dimensional representation $\bm{z}$, which is cached for test-time retrieval. Then at test time, the model outputs the representation of the input example $\bm{x}$, uses it as the query to search for $k$ nearest training instances in the cache store. 
In the meantime, the distance between the query and each retrieved training point can be calculated, and the corresponding ground-truth labels for the retrieved training points can also be extracted. The normalized distances are treated as the weight associated with each training point the model retrieves to interpret its prediction for the test input example, and the percentage of nearest neighbors belonging to the predicted class, which is called the {\it conformity score} \cite{wallace2018interpreting}, can be interpreted as how much the training data supports a classification decision. \cite{rajani2020explaining} uses the conformity score to calibrate an uncertain model prediction, showing that KNN interpretation is able to identify mislabeled examples. Figure \ref{fig:knn_kda} (left) provides an overview of the KNN-based interpretation method proposed in \cite{rajani2020explaining}.

KNN based interpretation works after the training process completes, and thus belongs to the {\it post-hoc} category.

\subsection{Landmark Based Interpretation}
\label{kernel}
Landmark based interpreting methods \cite{croce-etal-2018-explaining,croce2019auditing} use {\it landmarks}, which are a set of real reference training examples, to compile the linguistic properties of an unseen test example. This line of methods combine the Layerwise Relevance Propagation (LRP) \cite{bach2015lrp} and Kernel-based Deep Architectures (KDAs) \cite{croce-etal-2017-deep} to complete interpretation. 
An illustration is shown in Figure \ref{fig:knn_kda} (right).

More formally, given an input example $x$ and $d$ landmarks $\{l_1,l_2,\cdots,l_d\}$ sampled from the training set, the model first computes the similarity score $K(x,l_i)$ between $x$ and each landmark $l_i$ using a kernel function $K(\cdot,\cdot)$. Then, a Nystr{\"o}m layer \cite{NIPS2000_19de10ad} is used to map the vector of similarity scores $[K(x,l_1),K(x,l_2),\dots,K(x,l_d)]$ to a high-dimensional representation. Passing through multiple ordinary hidden layers, the model last makes its prediction at the classification layer. The expected explanation is obtained from the network output by applying LRP to revert the propagation process, therefore linking the output back to the similarity vector. Once LRP assigns a score to each of these landmarks (i.e., the corresponding similarity score $K(x,l_i)$), we can select the {\it positively active} landmarks as evidence in favor of a class $C$. A
template-based 
 natural language explanation 
 is generated 
  based on the activated landmarks. As an example given in \cite{croce2019auditing}, the generated explanation to the input example ``{\it What is the capital of Zimbabwe?}'' that refers to \texttt{Location} is ``{\it since it recalls me of} `{\it What is the capital of California}?', {\it which also refers to} \texttt{Location}''. In this example, ``{\it What is the capital of California?}'' is the activated landmark used to trigger the template-based explanation, and \texttt{Location} is the class $C$ the model predicts.

To decide whether the landmark--based model belonging to the joint or post-hoc category, 
we need to take into consideration 
its two constituent components: (1) the kernel architecture component which involves the kernel function $K(\cdot,\cdot)$ and the Nystr{\"o}m layer, and (2) the LRP scoring mechanism. For the former, the kernel architecture is trained together with the entire model and therefore the landmark--based model can be viewed as along  the {\it joint} interpretation line. For the latter, when computing the importance of each landmark,  LRP is a typical {\it post-hoc} interpreting method. 
The landmark--based model can thus be viewed as both joint and post-hoc.

\section{Test-Based Interpretation}
\label{test}
\begin{figure}[t]
  \centering
  \includegraphics[width=1\textwidth]{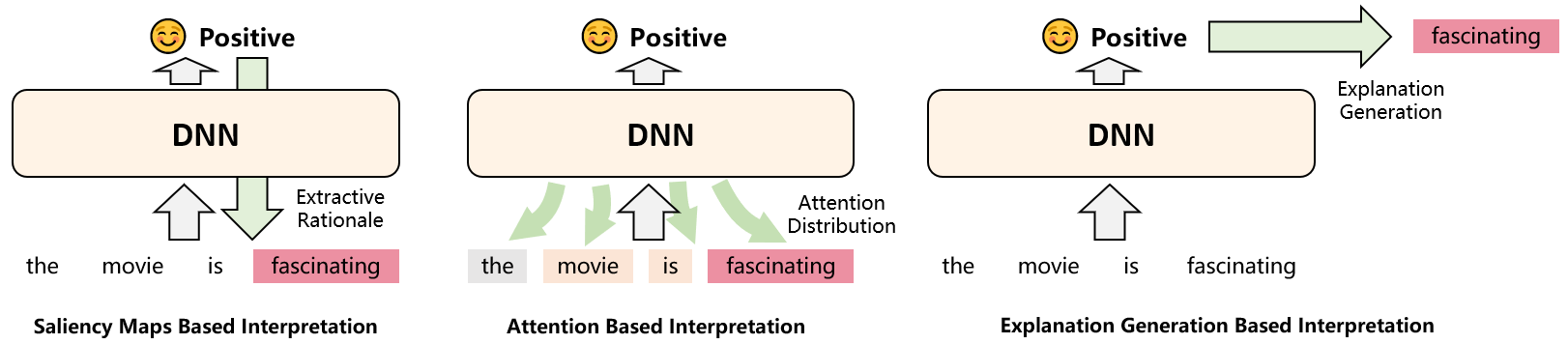}
  \caption{An overview of test-based methods. {\it Left}: The saliency score is the importance or relevance of a token or a span with respect to the model prediction. A higher saliency score denotes greater importance. {\it Middle}:  Attention scores over input tokens reflects how the model distributes its attention to different parts of the input. Attention distributions can thus be viewed as a tool for interpretation. {\it Right}: Given the input, the model gives its prediction as well as the evidence supporting its decision. A judicious decision can be explained in a way of generating the corresponding explanations.}
  \label{fig:test-overview}
\end{figure}

Different from training-based methods that focus on interpreting neural NLP models by identifying the training points responsible for the model's prediction on a specific test input, the test-based methods, instead, aim at providing explanations about which part(s) of the test example contribute most to the model prediction. This line of methods are rooted in the intuition that not every token or span within the input sentence contributes equally to the prediction decision, and thus the most contributive token or span can usually interpret the model behaviors. For example in Figure \ref{fig:overview}, the input test example  ``{\it the movie is fascinating}'' has a very typical adjective ``{\it fascinating}'' that would nudge the model to predict the class label \texttt{Positive}. In this sense, the most contributive part is word ``{\it fascinating}''. If we insert the negation ``{\it not}'' before ``{\it fascinating}'', the most contributive part would therefore be ``{\it not fascinating}'', a span rather than a word in the input sentence.

Most of existing test-based methods fall into the following three categories: saliency maps, attention as explanation and explanation generation.
Saliency maps and attention provide very straightforward explanations in the form of visualized heatmaps distributed to the input text, showing intuitive understandings of which part(s) of the input the model focuses most on, and which part(s) of the input the model pays less attention to. The methods of textual explanation generation justify the model prediction by generating causal evidence either from a part of the input sentence, from external knowledge resources or completely from scratch.
This test-based framework of model interpretation, in contrast to training-based methods, is concerned with one particular input example, and conveys useful and intuitive information to end-users even without any AI background.

Figure \ref{fig:test-overview} shows a high-level overview for the three different strands of test-based methods.
In the rest of this subsection, we will describe each line of methods in detail.

\subsection{Saliency-based Interpretation}
\label{saliency}
\begin{table}[t]
    \centering
    \begin{tabular}{cc}
        \toprule 
        {\bf Method} & {\bf Formulation}\\\midrule
        Vanilla Gradient & $\text{Score}(x_i)=\frac{\partial S_y(x)}{\partial x_i}$ \\
        Integrated Gradient & $\text{Score}(x_i)=(x_i-\bar{x}_i)\cdot\int_{\alpha=0}^1\frac{\partial S_y(\tilde{x})}{\partial \tilde{x}_i}\big\vert_{\tilde{x}=\bar{x}+\alpha(x-\bar{x})}\mathrm{d}\alpha$\\
        Perturbation & $\text{Score}(x_i)=S_y(x)-S_y(x\backslash\{x_i\})$\\
        LRP & $\text{Score}(x_i)=r^{(0)}_i,~r^{(l)}_i=\sum_j\frac{z_{ji}}{\sum_{i'}(z_{ji'}+b_j)+\epsilon\cdot\text{sign}(\sum_{i'}(z_{ji'}+b_j))}r^{(l+1)}_j$\\
        DeepLIFT & $\text{Score}(x_i)=r^{(0)}_i,~r^{(l)}_i=\sum_j\frac{z_{ji}-\bar{z}_{ji}}{\sum_{i'}z_{ji'}-\sum_{i'}\bar{z}_{ji'}}r^{(l+1)}_j$\\
        \bottomrule
    \end{tabular}
    \caption{Mathematical formulation of different saliency methods. LRP and DeepLIFT perform in a top-down recursive manner.}
    \label{tab:saliency-overview}
\end{table}

In natural languages, some words are more {\it important} than other words in indicating the direction to which the model predicts. In sentiment classification for example, the words associated with strong emotions such as ``{\it fantastic}'', ``{\it scary}'' and ``{\it exciting}'' would provide interpretable cues for the model prediction. If we could measure the {\it importance} of a word, or a span, and examine the relationship of the most important part(s) with the model prediction, we will be able to interpret the model behaviors through the lens of these important contents and their relationship with model predictions. The {\it importance} of each token (or its corresponding input feature), which sometimes is also called the {\it saliency} or the {\it attribution}, defines the relevance of that token with the model prediction. By visualizing the {\it saliency maps}, i.e., plotting the saliency scores in the form of heatmaps, users can easily understand why model make decisions by examining the most salient part(s).

We describe four typical ways of computing the saliency score in the rest of this subsection. These methods include: gradient-based, perturbation-based, LRP-based and DeepLIFT-based. Table \ref{tab:saliency-overview} summarizes the mathematical formulations of these methods.

\subsubsection{Gradient-based saliency scores}
Gradient-based saliency methods compute the importance of a specific input feature (e.g., vector dimension, word or span) based on the first-order derivative with respect to that feature. 
Suppose that $S_y(x_i)$ is output with respect to class label $y$ before applying the softmax function, and $x_i$ is the input feature, which is 
the $i$th dimension of 
 the word embedding  $x = \{x_1, x_2, ..., x_i, ..., x_N\}$ for word $w$
in NLP tasks. We have:
\begin{equation}
S_y({x_i} + \Delta {x_i})\approx S_y({x_i}) + \frac{\partial S_y(x_i)}{\partial x} \Delta {x_i}
\end{equation}
$\frac{\partial S_y(x_i)}{\partial x_i}$ can be viewed as the change of $S_y({x_i})$ with respect to 
the change of
$x_i$. 
If $\frac{\partial S_y(x_i)}{\partial x_i}$  approaches 0, it means that $S_y({x_i})$ is not at all sensitive to the change of $x_i$.
$\frac{\partial S_y(x_i)}{\partial x}$  can thus be straightforwardly viewed as the importance of the word dimension $x_i$  \cite{li2015visualizing}:
\begin{equation}
  \text{Score}(x_i)=\frac{\partial S_y(x)}{\partial x_i}
  \label{eq:gradient}
\end{equation}
Equation \ref{eq:gradient} computes the importance of a single dimension for the word vector. 
To measure the importance of a word $\text{Score}(w)$, we can use the norm as follows:
\begin{equation}
\text{Score}(w)=\sqrt{\sum_i \left\Vert\frac{\partial S_y(x_i)}{\partial x_i}\right\Vert^2}
\end{equation}
or multiply the important vector with the input feature vector \cite{denil2015extraction,ancona2017towards}:
\begin{equation}
\text{Score}(w)=\left\vert\sum_i x_i\frac{\partial S_y(x_i)}{\partial x_i}\right\vert
\end{equation}
Both methods work well for the interpretation purpose.
Albeit simple and easy to implement, this kind of vanilla gradient saliency score assumes a linear approximation of the relationship between the input feature and the model output, which is not the case for deep neural networks. \cite{sundararajan2017axiomatic} proposed  the {\it integrated gradient (IG)}, a modification to the vanilla gradient saliency score.
IG computes the average gradient 
along the linear path of  varying the input from a baseline value $\bar{x}$ to itself $x$, which produces a more reliable and stable result compared to the vanilla gradient approach. 
 The baseline value is often set to zero. IG can be formulated as the following equation:
\begin{equation}
  \text{Score}(x_i)=(x_i-\bar{x}_i)\cdot\int_{\alpha=0}^1\frac{\partial S_y(\tilde{x})}{\partial \tilde{x}_i}\big\vert_{\tilde{x}=\bar{x}+\alpha(x-\bar{x})}\mathrm{d}\alpha
  \label{eq:ig}
\end{equation}

\subsubsection{Perturbation-based saliency scores}
\begin{figure}[t]
    \centering
    \begin{minipage}[t]{0.48\textwidth}
      \centering
      \includegraphics[width=1\textwidth]{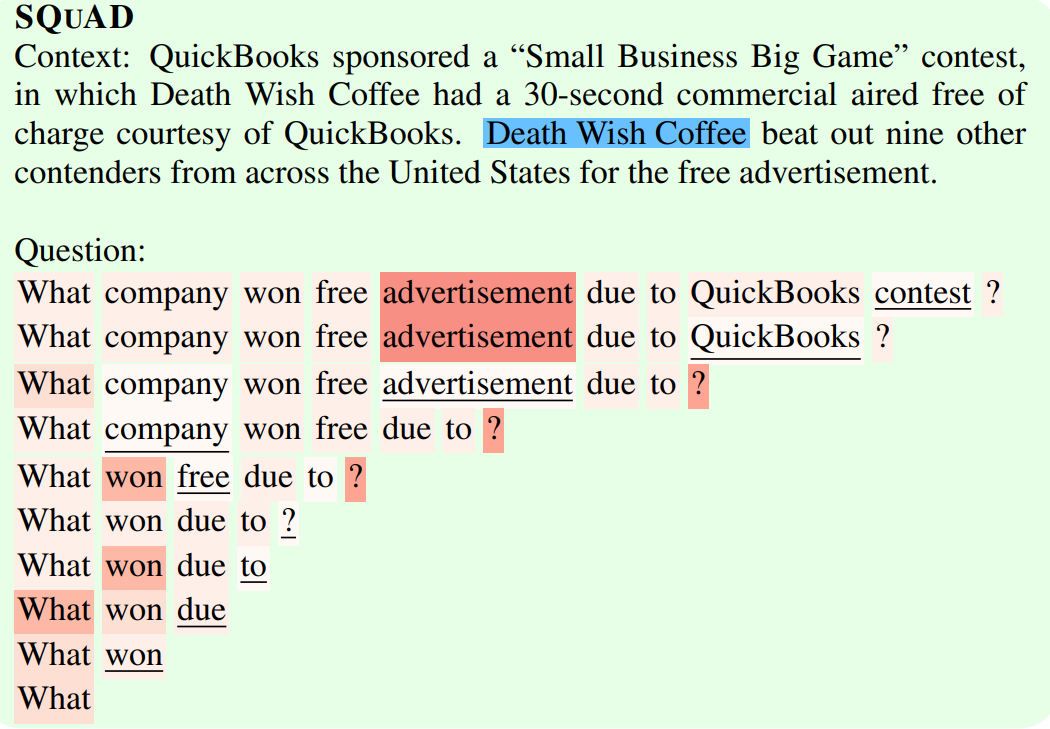}
    \end{minipage}%
    \hfill
    \begin{minipage}[t]{0.48\textwidth}
      \centering
      \includegraphics[width=1\textwidth]{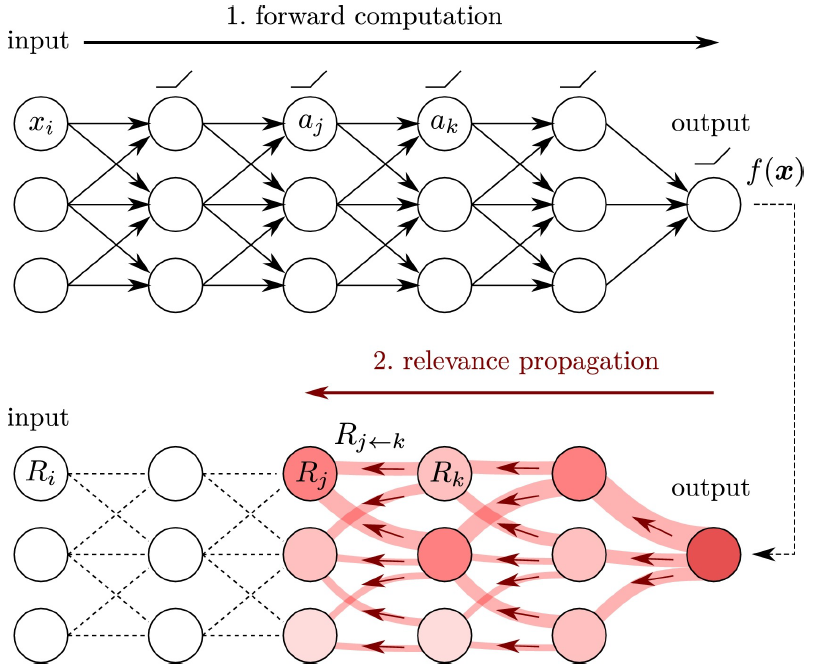}
    \end{minipage}
    \caption{Diagram of the perturbation based method (left, the figure is brought from \cite{feng2018pathologies}) and LRP (right, the figure is brought from \cite{gholizadeh2021model}). {\it Left}: At each step, the least important word is removed by the saliency score using the leave-one-out technique. As shown in the figure, the heatmaps generated with leave-one-out shift drastically: the word ``advertisement'' is the most salient in step one and two, but becomes the least salient in step three. {\it Right}: LRP provides decomposition of the attribute scores from the model prediction to the input features. It does not utilize partial derivatives but rather directly decomposes the activation function to produce the attribute score.}
    \label{fig:path-lrp}
\end{figure}

Perturbation-based methods compute the saliency score of an input feature by removing, masking or altering that feature, passing the altered input again into the model and measuring the output change. This technique is straightforward to understand: if a particular input token is  important, then removing it  will be more likely to causing drastic prediction change and flip the model prediction.

One simple way to perturb an input is to erase a word from the input, and examine the change in the model's prediction on the target class label \cite{li2016understanding}. This technique is called the {\it leave-one-out} perturbation method. By iterating over all words, we can find the most salient word that leads to the largest prediction change. The saliency score using leave-one-out can be given as follows:
\begin{equation}
  \text{Score}(x_i)=S_y(x)-S_y(x\backslash\{x_i\})
  \label{eq:leave-one-out}
\end{equation}
The token-level removal can be extended to the span level or the phrase level \cite{rosa2019inducing,wu2020perturbed}.
Based on the leave-one-out technique, \cite{feng2018pathologies} uses input reduction, which iteratively removes the least salient word at a time, to induce the {\it pathological behaviors} of neural NLP models. An example is shown in Figure \ref{fig:path-lrp} (left).

Another way to perturb the input is to inject learnable {\it interpretable adversaries} into an input feature \cite{sato2018interpretable,guan2019towards}. By conforming to specific restrictions (e.g., the direction, the magnitude or the distribution) and learning to optimize predefined training objectives, the adversarial perturbations are able to reveal some interpretable patterns within neural NLP models, such as the word relationships, the information of each word discarded when words pass through layers, and how the model evolves during training.

\subsubsection{LRP-based saliency scores}
\label{sec:lrp}
Gradient-based and perturbation-based saliency methods directly compute influence of input layer features on the output in an end-to-end fashion, 
 and thus neglects
 the dynamic changes of word saliency in intermediate layers. Layerwise Relevance Propagation \cite{layer-relev-propagation}, or LRP for short, provides full, 
 layer-by-layer
  decomposition for the attribute values from the model prediction backward to the input features in a recursive fashion. A sketch of the propagation flow of LRP is shown in Figure \ref{fig:path-lrp} (right).

LRP has been applied to interpret neural NLP models for various downstream tasks \cite{arras2016explaining,arras2017relevant,arras2017explaining,ding2017visualizing}. LRP defines a quantify $r^{(l)}_i$ -- the ``relevance'' of neuron $i$ in layer $l$, and starts relevance decomposition from the output layer $L$ to the input layer $0$, redistributing each relevance $r^{(l)}_i$ down to all the neurons $r^{(l-1)}_j$ at its successor layer $l-1$. The relevance redistribution across layers goes as follows:
\begin{equation}
  \begin{aligned}
    r^{(L)}_i&=
    \begin{cases}
      S_y(x),&i~\text{is the target unit of the gold class label}\\
      0,&\text{otherwise}
    \end{cases}\\
    r^{(l)}_i&=\sum_j\frac{z_{ji}}{\sum_{i'}(z_{ji'}+b_j)+\epsilon
    \cdot\text{sign}(\sum_{i'}(z_{ji'}+b_j))}r^{(l+1)}_j
    \label{eq:lrp}
  \end{aligned}
\end{equation}
where $z_{ji}=w^{(l+1,l)}_{ji}x^{(l)}_i$ is the weighted activation of neuron $i$ in layer $l$ onto neuron $j$ in layer $l+1$ and $b_j$ is the bias term. A small $\epsilon$ is used to circumvent numerical instabilities. The saliency scores at the input layer can thus be defined as $\text{Score}(x_i)=r^{(0)}_i$. Equation \ref{eq:lrp} reveals that the more relevant an input feature is to the model prediction, the larger relevance (saliency) score it will gain backward from the output layer. A merit LRP offers is that it allows users to understand how the relevance of each input feature flows and changes across layers.
This  property enables interpretation from the inside of model rather than the superficial operation of 
associating input features with the output.

\subsubsection{DeepLIFT-based saliency scores}
\begin{figure}[t]
  \centering
  \includegraphics[width=1\textwidth]{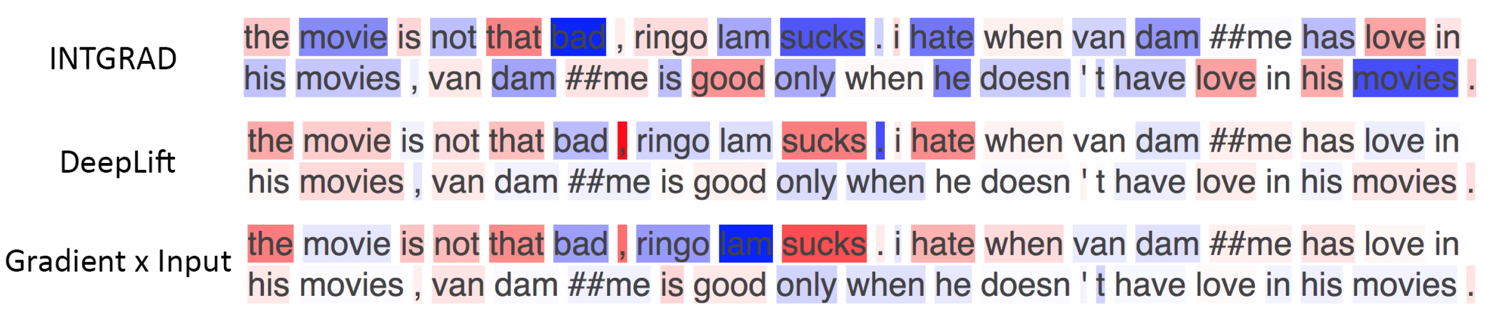}
  \caption{A visualization of saliency methods Integrated Gradient, DeepLIFT and Gradient$\times$Input (the figure is  borrowed from \cite{bodria2021benchmarking}). Red denotes a positive saliency value while blue denotes negative values, and the greater the absolute value is, the darker the color would be.}
  \label{fig:test-saliency}
\end{figure}

Similar to LRP, DeepLIFT \cite{shrikumar2017learning,arkhangelskaia2019whatcha}  proceeds in a progressive backward fashion, but the relevance score is calculated in a relative view: instead of directly propagating and redistributing the relevance score from the upper layer to the lower layer, DeepLIFT assigns each neuron a score that represents the relative effect of the neuron at the original input $x$ relative to some reference input $\bar{x}$, taking the advantages of  both Integrated Gradient and LRP. The mathematical formulation of DeepLIFT can be expressed as follows:
\begin{equation}
  \begin{aligned}
    r^{(L)}_i&=
    \begin{cases}
      S_y(x)-S_y(\bar{x}),&i~\text{is the target unit of the gold class label}\\
      0,&\text{otherwise}
    \end{cases}\\
    r^{(l)}_i&=\sum_j\frac{z_{ji}-\bar{z}_{ji}}{\sum_{i'}z_{ji'}-\sum_{i'}\bar{z}_{ji'}}r^{(l+1)}_j
    \label{eq:deeplift}
  \end{aligned}
\end{equation}
Using the baseline input $\bar{x}$, the reference relevance values $\bar{z}_{ji}$ can be computed by running with $\bar{x}$ a forward pass through the neural model. Ordinarily, the baseline is set to be zero. $z_{ji}$ and $\bar{z}_{ji}$ are computed in the same way but use different inputs: $z_{ji}=w^{(l+1,l)}_{ji}x^{(l)}_i$ and $\bar{z}_{ji}=w^{(l+1,l)}_{ji}\bar{x}^{(l)}_i$.

Figure \ref{fig:test-saliency} shows the saliency maps generated by three saliency methods: integrated gradient (IG), DeepLIFT and Gradient$\times$Input, with respect to the input sentence ``{\it the movie is not that bad, ringo lam sucks. i hate when van dam \#\#me has love in his movies, van dam \#\#me is good only when he doesn ' t have love in this movies.}'' 
As can be seen from the figure, the highlighted words are very different across different methods. IG produces meaningful explanations as it correctly marks out the words  ``bad'', ``sucks'' and ``hate'' that strongly guide the model prediction toward negative, whereas DeepLIFT and vanilla gradient struggle to correctly detect the interpretable units. This is a case where IG works well while DeepLIFT and vanilla gradient do not, while in other cases, the superiority of IG is not guaranteed. 
We thus should take multiple interpreting methods for full consideration. 

Because saliency-based interpretation uses an independent probing model to interpret the behaviors of the main model, the methods  belong to the {\it post-hoc} category.

\subsection{Attention-based Interpretation}
\subsubsection{Background}
\label{attention}
\begin{figure}[t]
  \centering
  \includegraphics[width=1\textwidth]{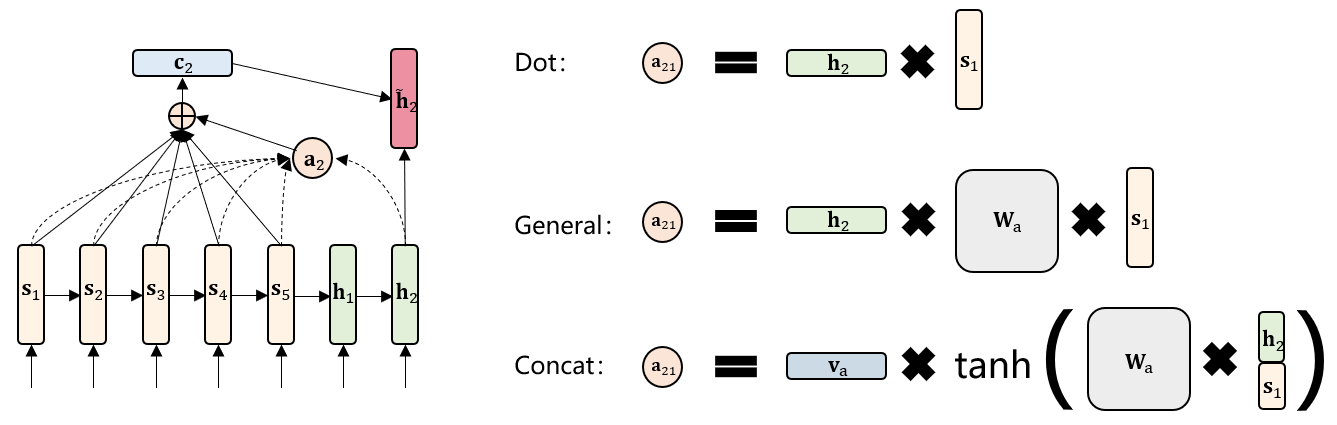}
  \caption{A visualization of the attention mechanism. We show the three forms of attention proposed in \cite{luong-etal-2015-effective}.}
  \label{fig:attention}
\end{figure}

When humans read, 
they only pay {\it attention} to the key part. For example, 
to answer  the question ``What causes precipitation to fall'',
given the document context ``In meteorology, precipitation is any product of the condensation of atmospheric water vapor that falls under gravity.'' 
 humans could easily identify the correct phrase ``falls under gravity'' in which the answer ``gravity'' resides, and other parts of the document are less attended.

The attention mechanism, grounded on the way how humans read,  has been widely used in  NLP \cite{bahdanau2014neural,luong-etal-2015-effective}. 
 Take the neural machine translation task as an example. Assume the hidden representations of the input document is $s=\{s_1,s_2,\cdots,s_n\}$, where each $s_i$ is a $d$-dimensional vector and $n$ is the input length, and the decoder sequentially generates one token per time step:
\begin{equation}
  p(y_j|y_{<j},s)=\text{softmax}(W\tilde{h}_j),~\tilde{h}_j=\text{Attention}(h_j,s),~h_j=f(h_{j-1})
  \label{eq:attention}
\end{equation}
$W$ is a transformation matrix that outputs a vocabulary-size vector, $f$ computes the current hidden state given the previous state and can be of any recurrent structure, and \texttt{Attention} is the attention component that takes as input the current hidden state and the source hidden representations and outputs a new attended hidden state, which is then fed to the softmax layer for model prediction. The core idea of attention is to derive a context vector $c_j$ that captures the weighted source hidden representations, and the weight notifies how much should the current time step should pay attention to each of the source input token. This can be formalized as:
\begin{equation}
  a_{ji}=\text{align}(h_j,s_i)=\frac{\exp(\text{score}(h_j,s_i))}{\sum_{i'}\exp(\text{score}(h_j,s_{i'}))}
  \label{eq:attention-weight}
\end{equation}
Once obtaining the weight $a_{ji}$, the aggregated hidden state $\tilde{h}_j$ in Equation \ref{eq:attention} can be calculate by:
\begin{equation}
  \tilde{h}_j=\tanh(W_c[c_j;h_j]),~c_j=\sum_{i=1}^n a_{ji}s_i
\end{equation}
The remaining problem is how to compute the attention weight $a_{ji}$ in Equation \ref{eq:attention-weight}. \cite{luong-etal-2015-effective} proposed three forms of scores: the {\it dot} form, the {\it general} form, and the {\it concat} form. We provide a figure illustration in Figure \ref{fig:attention}. They can be formalized as follows:
\begin{equation}
  \text{score}(h_j,s_i)=
  \begin{cases}
    h_j^\top s_i,&\text{dot}\\
    h_j^\top W_a s_i,& \text{general}\\
    v_a^\top\tanh(W_a[h_j;s_i]),&\text{concat}
  \end{cases}
\end{equation}
\cite{vaswani2017transformer} proposed {\it self-attention}, where attention is performed between different positions of a {\it single} sequence rather than two separate sequences in order to compute a representation of the sequence. The expression of self-attention is very like to the dot form of \cite{luong-etal-2015-effective}, but with differences that each token has three representations -- the query $q$, the key $k$ and the value $v$, and a scaled factor $\sqrt{d_k}$ is used to avoid overlarge numbers:
\begin{equation}
  \text{SelfAttention}(q_j,k_i,v_i)=\frac{\exp(q_j^\top k_i/\sqrt{d_k})}{\sum_{i'}\exp(q_j^\top k_{i'}/\sqrt{d_k})}v_i
\end{equation}
More importantly, self-attention works in conjunction with the {\it multi-head} mechanism. Each head represents a sub-network and runs self-attention independently. The produces results are then collected and processed as input to the next layer.

Regardless of different forms of attention, the core idea is the same:  {\it paying attention to the most crucial part in the document}. 
Given attention weights, we can regard the weight as a measure of importance the model assigns to each part of the input.
The following  questions arise regarding model interpretation through attention weights: 
(1) can the learned attention patterns indeed capture {\it the most crucial} context? (2) 
How the attention mechanism can be leveraged to interpret model decisions? (3)Does attention reflect some linguistic knowledge when it is applied to neural models? This section describes  typical works that target the above three questions.

\subsubsection{Is attention interpretable?}
There has been a growing debate on the question of {\it whether attention is interpretable} \cite{serrano2019attention,pruthi2019learning,jain2019attention,wiegreffe-pinter-2019-attention,vashishth2019attention,Brunner2020On}. The 
debate centers around
whether higher attention weights denote more influence on the the model prediction, which is sometimes unclear. 

\cite{serrano2019attention,pruthi2019learning,jain2019attention,Brunner2020On} hold negative attitude towards attention's interpretability. \cite{jain2019attention} empirically assessed the degree to which attention weights provide meaningful explanations for model predictions. They raised two properties attention weights should hold
if they provide faithful explanations: (1) attention weights should correlate with feature importance-based methods such as gradient-based methods (Section \ref{saliency}); (2) altering attention weights ought to yield corresponding changes in model predictions. However, through extensive experiments in the context of text classification, question answer and natural language inference using a BiLSTM model with a standard attention mechanism, they found neither property is consistently satisfied: for the first property, they observed that attention weights do not provide strong agreements with the gradient-based approach \cite{li2015visualizing} or the leave-one-out approach \cite{li2016understanding}, as measured by the Kendall $\tau$ correlation; for the second property, by randomly re-assigning attention weights or generating adversarial attention distributions, \cite{jain2019attention} found that alternative attention weights do not essentially change the model outputs, indicating that the learned attention patterns cannot consistently provide transparent and faithful explanations.
\cite{serrano2019attention} reached a similar conclusion to \cite{jain2019attention} by examining intermediate representations to assess whether attention weights can provide consistent explanations. They used two ways to measure attention interpretability: (1)  calculating the difference between two JS divergences -- one coming from the model's output distribution after zeroing out a set of the most attended tokens, and the other coming from the distribution after zeroing out a random set of tokens; (2)  computing the fraction of decision flips caused by zeroing out the most attended tokens. Experiments demonstrate that higher attention weights has a larger impact on neither of these two measures.

\begin{figure}[t]
  \centering
  \includegraphics[width=0.8\textwidth]{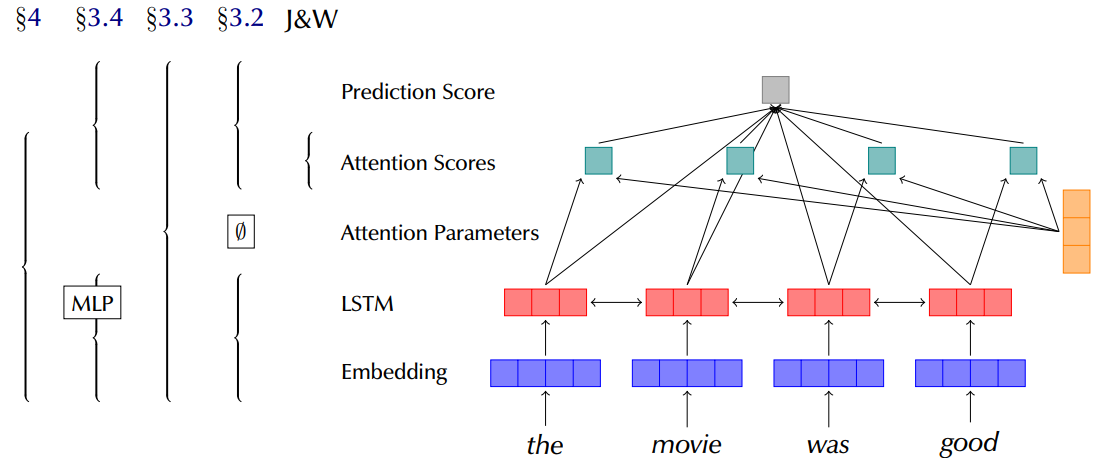}
  \caption{A schematic diagram of an LSTM model with attention (the figure is borrowed from \cite{wiegreffe-pinter-2019-attention}). \cite{jain2019attention} only manipulated attention at the {\it attention scores} level, while \cite{wiegreffe-pinter-2019-attention} manipulated attention at various levels and components of the model.}
  \label{fig:attention-wiegreffe}
\end{figure}

Positive opinions come from \cite{wiegreffe-pinter-2019-attention,vashishth2019attention}. \cite{wiegreffe-pinter-2019-attention} challenged the opinion of \cite{jain2019attention} and argued that testing the attention's interpretability needs to take into account all elements of the model, rather than solely the attention scores. To this end, \cite{wiegreffe-pinter-2019-attention} proposed to assess attention's interpretability  from four perspectives (an illustration is shown in Figure \ref{fig:attention-wiegreffe}): (1) freeze the attention weights to be a uniform distribution and find that this variant performs as well as learned attention weights, indicating that the adversarial attention distributions proposed in \cite{jain2019attention} are not evidence against attention as explanation in such cases; (2) examine the expected variance induced by multiple training runs with different initialization seeds and find that the attention distributions seem not to be distanced much; (3) use a simple diagnostic tool which tests attention distributions by applying them as frozen weights in a non-contextual multi-layer perceptron model, and find that the model achieves better performance compared to a self-learned counterpart, demonstrating that attention indeed provides meaning model-agnostic explanations; (4) propose a model-consistent training protocol to produce adversarial attention weights and find that the resulting adversarial attention does not perform well in the diagnostic MLP setting. These findings correct some of the flaws in \cite{jain2019attention} and suggest that attention can provide plausible and faithful explanations. 

\cite{vashishth2019attention} tested attention interpretability on a bulk of NLP tasks including text classification, pairwise text classification and text generation. Besides vanilla attention, self-attention is also considered for comparison. Through automatic evaluation and human evaluation, \cite{vashishth2019attention} concluded that for tasks involving two sequences, attention weights are consistent with feature importance methods and human evaluation; but for single sequence tasks, attention weights do not provide faithful explanations regarding the model performance. 

Upon the time of this review, whether the attention mechanism can be straightforwardly used for model interpretation is still debatable. 

\subsubsection{Attention for interpreting model decisions}
\begin{figure}[t]
    \centering
    \begin{minipage}[t]{0.4\textwidth}
      \centering
      \includegraphics[width=1\textwidth]{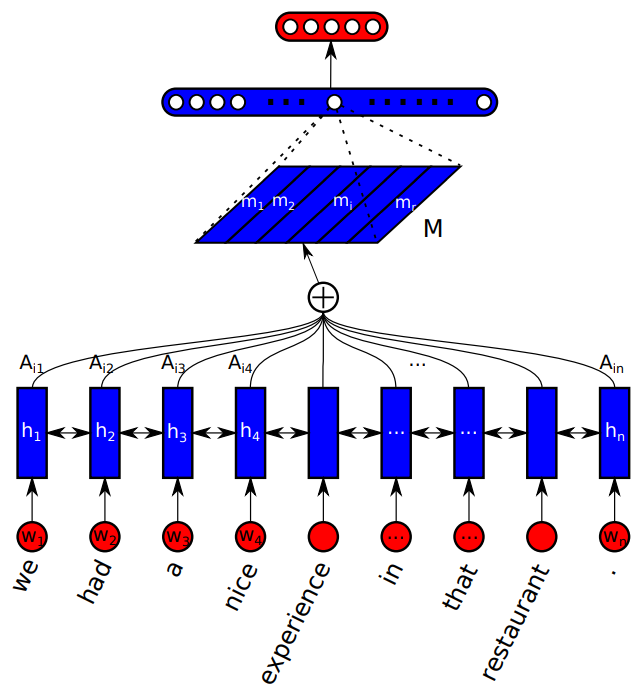}
    \end{minipage}%
    ~~~~~~~~~
    \begin{minipage}[t]{0.3\textwidth}
      \centering
      \includegraphics[width=1\textwidth]{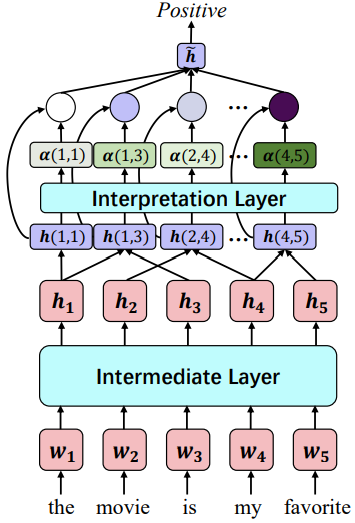}
    \end{minipage}
    \caption{Diagrams of the self-attentive sentence embedding model \cite{wang2016attention} and attention-based self-explaining structure \cite{sun2020selfexplaining}. The figures are taken from the original literature.}
    \label{fig:attention-self}
\end{figure}
Putting aside the ongoing debate,
here we describe recently proposed attention-based interpretation methods. 
An early work comes from \cite{wang2016attention}, who learned aspect embeddings to guide the model to make aspect-specific decisions. When integrated with different aspects, the learned attention shows different patterns for an input sentence. As an example shown in Figure 5 of the original literature, given the input ``the appetizers are ok, but the service is slow'', when the \texttt{service} aspect embedding is integrated, the model attends to the content of ``service is slow'', and when the \texttt{food} aspect embedding is integrated, the model attends to ``appetizers are ok'', displaying that the attention can, at least to some degree, learn the most essential part of the input according to different aspects of interest.

For neural machine translation, 
during each decoding step, the model needs to attend to the most relevant part of the source sentence, and then translates it into the target language \cite{bahdanau2014neural,luong-etal-2015-effective}.
\cite{ghader-monz-2017-attention} investigated the word alignment induced by attention and compared the similarity of the induced word alignment to traditional word alignment. They concluded that attention agrees with traditional alignment to some extent: consistent in some cases while not in other cases.
\cite{lee2017interactive} built an interactive visualization tool for manipulating attention in neural machine translation systems. This tool supports automatic and custom attention weights and visualizes output probabilities in accordance with the attention weights. This helps users to understand how attention affects model predictions.
For natural language inference, 
\cite{ghaeini2018interpreting} examined the attention weights and the attention saliency, illustrating that attention heatmaps  identify the alignment evidence supporting model predictions, and the attention saliency shows how much such alignment impacts the decisions.

\cite{lin2017structured,sun2020selfexplaining} proposed self-interpreting structures that can improve model performance and interpretability at the same time. The key idea is to learn an attention distribution over the input itself (and may be at different granularity, e.g., the word level or the span level). Such attention distribution would be interpreted to some extend as evidence of how the model reasons about the given input with respect to the output. The core idea behind \cite{lin2017structured} is to learn an attention distribution over the input word-level features and average the input features according to the attention weights. They further extended the standalone attention distribution to multiple ones, each focusing on a different aspect and leading to a specific sentence embedding. This enables direct interpretation from the learned sentences embeddings. Suppose the input features are represented as a feature matrix $H\in\mathbb{R}^{n\times d_1}$ where $n$ is input length and $d_1$ is the feature dimensionality. Two matrices $W_1\in\mathbb{R}^{d_2\times d_1}$ and $W_2\in\mathbb{R}^{r\times d_2}$ are used to transform the feature matrix into attention logits, followed by a softmax operator to derive the final attention weights:
\begin{equation}
  A=\text{softmax}(W_2\tanh(W_1H^\top))
\end{equation}
The resulting attention matrix $A$ is of size $r\times n$, where each row is a unique attention distribution adhering to a particular aspect. Finally, the sentence embeddings $M\in\mathbb{R}^{r\times d_1}$ are obtained by applying $A$ to the original feature matrix $H$:
\begin{equation}
  M=AH
\end{equation}
According to different output classes, the self-attentive model is able to capture different parts of the input responsible for the model prediction.

\cite{sun2020selfexplaining} extended this idea to the span-level, rather than just the word-level. They first represent each span by considering the start-index representation and the end-index representation of that span, forming a span-specific representation $h(i,j)$; then they treat these spans as basic units and compute an attention distribution in a way similar to \cite{lin2017structured}; last, the spans are averaged according to the learned attention weights for final model decision. This span-level strategy provides better flexibility to model interpretation.
An overview of the self-interpreting structures proposed in \cite{lin2017structured} and \cite{sun2020selfexplaining} is shown in Figure \ref{fig:attention-self}.

\subsubsection{BERT attention encoding linguistic notions}
Built on top of the self-attention mechanism and Transformer blocks, BERT-style pretraining models  \cite{devlin2018bert,yang2019xlnet,danqi2020spanbert,liu-etal-2019-multi,Clark2020ELECTRA,performer,radford2019gpt2,brown2020language,lewis2019bart,song2019mass,NEURIPS2019_c20bb2d9,sun2021chinesebert}
have been used as the backbone
 across a wide range of downstream NLP tasks. The attention pattern behind BERT
 encodes  meaning linguistic knowledge \cite{rogers2020primer}, which cannot be easily observed in LSTM based models. 

\begin{figure}[t]
  \centering
  \includegraphics[width=1\textwidth]{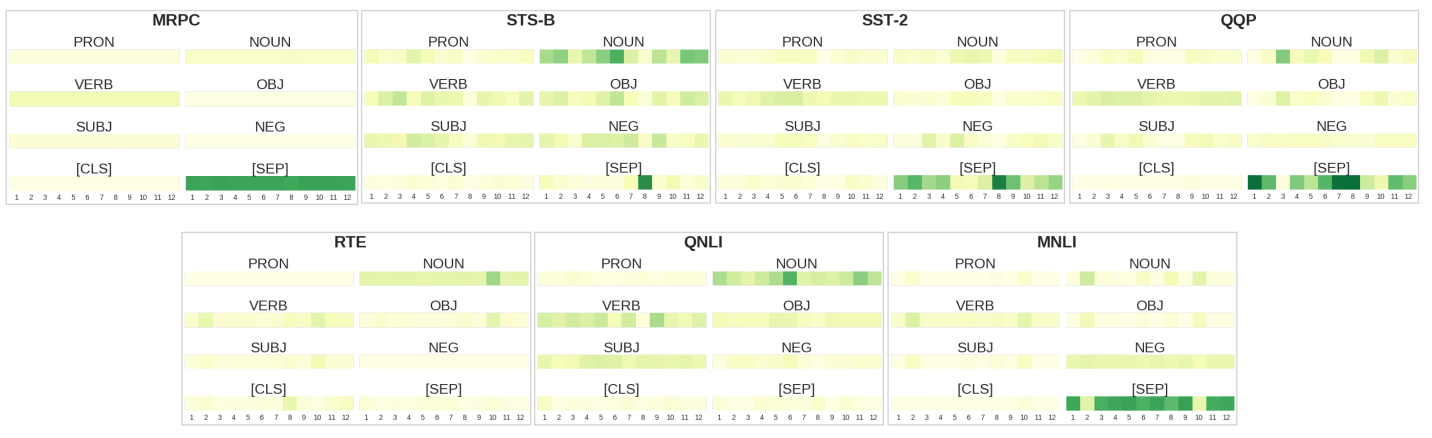}
  \caption{Per-task attention weights to part-of-speeches and the special \texttt{[SEP]} token averaged over input length and over the entire dataset. Except the \texttt{[SEP]}, noun words are the most focused (the figure is brought from \cite{kovaleva-etal-2019-revealing}).}
  \label{fig:attention-pos}
\end{figure}

\paragraph{BERT attention reveals part-of-speech patterns}~{}\\
\cite{vig2019analyzing} observed that BERT encodes   part-of-speech (POS) tags. 
POS is a category of words that have similar properties and display similar syntactic behaviors. Commonly, English part-of-speeches include noun, verb, adjective, adverb, pronoun, preposition, conjunction, etc. \cite{vig2019analyzing} observed that the attention heads that focus on a particular POS tag tend to cluster by layer depth. For example, the top heads attending to proper nouns are all in the last three layers of the model, which may be due to that deep layers focus more on named entities; the top heads attending to determiners are all in the first four layers of the model, indicating that deeper layers focus on higher-level linguistic properties. \cite{kovaleva-etal-2019-revealing} studied POS attention distributions on a variety of natural language understanding tasks. They found that the nouns are the most attended except the special \texttt{[SEP]} token used for classification (as shown in Figure \ref{fig:attention-pos}). These discoveries reveal that BERT can encode POS patterns through attention distributions, but may vary across layers and tasks. 

\begin{figure}[t]
  \centering
  \begin{minipage}[t]{0.6\textwidth}
    \centering
    \includegraphics[width=1\textwidth]{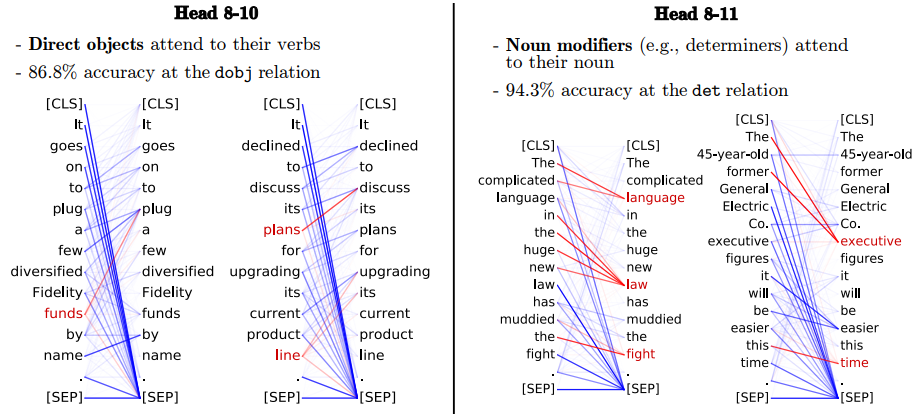}
  \end{minipage}%
  \begin{minipage}[t]{0.39\textwidth}
    \centering
    \includegraphics[width=1\textwidth]{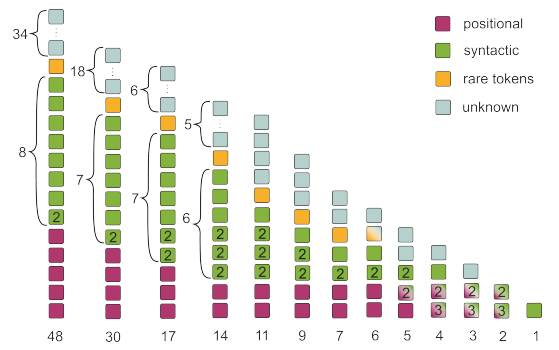}
  \end{minipage}
  \caption{{\it Left}: Two examples of BERT attention heads that encode dependency relations (the figure is brought from \cite{clark2019does}). {\it Right}: The functions of attention heads retrained after pruning (the figure is brought from \cite{voita-etal-2019-analyzing}).}
  \label{fig:attention-dep}
\end{figure}

\paragraph{BERT attention reveals dependency relation patterns}~{}\\
Dependency relation provides a principled way to structuralize a sentence's syntactic properties. A dependency relation between a pair of words is a head-dependent relation, where a labeled arc links from the head to the dependent, representing their syntactic relationship. Dependency relations depict the fundamental syntax information of a sentence and are crucial for understanding the semantics of that sentence. Extensive works show that BERT encodes dependency relations \cite{goldberg2019assessing,vig2019analyzing,clark2019does,htut2019attention,lin-etal-2019-open,voita-etal-2019-analyzing,reif2019visualizing}.
\cite{goldberg2019assessing} found that BERT consistently assigns higher attention scores to the correct verb forms as opposed to the incorrect one in a masked language modeling task, suggesting some ability to model subject-verb agreement.  Similar phenomena are observed by \cite{lin-etal-2019-open}. \cite{clark2019does,htut2019attention} decoupled the ``dependency relation encoding'' ability from different heads and concluded that there is no single attention head that does well at syntax across all types of relations, and that certain attention heads specialize to specific dependency relations, achieving high accuracy than fixed-offset baselines. Figure \ref{fig:attention-dep} (left) shows several examples of dependency relations encoded in specialized attention heads. \cite{voita-etal-2019-analyzing} investigated in depth the linguistic roles different heads play through attention patterns on the machine translation task. They found that only a small fraction of heads encode important linguistic features including dependency relations, and after pruning unimportant heads, the model still captures dependency relations in the rest heads and its performance does not significantly degrade. Figure \ref{fig:attention-dep} (right) shows the functions of heads retrained after pruning. \cite{reif2019visualizing} proposed {\it attention probe}, in which a {\it model-wide attention vector} is formed by concatenating the entries $a_{ij}$ in every attention matrix from every attention head in every layer. Feeding this vector into a linear model for dependency relation classification, we are able to understand whether the attention patterns indeed encode dependency information. They had an accuracy of 85.8\% for binary probe and an accuracy of 71.9\% for multi-class probe, indicating that  syntactic information is in fact encoded in the attention vectors.

\paragraph{BERT attention reveals negation scope}~{}\\
BERT encodes is the negation scope \cite{zhao-bethard-2020-berts}. {\it Negation} is a grammatical structure that reverses the truth value of a proposition. The tokens that express the presence of negation are the {\it negation cue} such as word ``no'', and the tokens that are covered in syntax by the negation cue are within the {\it negation scope}.
\cite{zhao-bethard-2020-berts} studied whether BERT pays attention to negation scope, i.e., the tokens within the negation scope are attending to the negation cue token. They found that before fine-tuning, several attention heads consistently encode negation scope knowledge, and outperform a fixed-offset baseline. After fine-tuning on a negation scope task, the average sensitivity of attention heads toward negation scope detection improves for all model variants. Their experiment results provide evidence for BERT's ability of encoding negation scope.

\paragraph{BERT attention reveals coreference patterns}~{}\\
BERT attention  encodes semantic patterns, such as patterns for coreference resolution. Coreference resolution is the task of finding all mentions that refer to the same entity in text. For example, in the sentence ``I voted for Nader because he was most aligned with my values'', the mentions ``I'' and ``my'' refer to the same entity, which is the speaker, and the mentions ``Nader'' and ``he'' both refer to the person ``Nader''. Coreference resolution is a challenging semantic task because it usually requires longer semantic dependencies than syntactic relations. \cite{clark2019does} computed the percentage of the times the head word of a coreferent mention attends to the head of one that mention's antecedents.  They find that attention achieves decent performances, attaining an improvement by over 10 accuracy points compared to a string-matching baseline and performing on par with a rule-based model. \cite{lin-etal-2019-open} explored the reflexive anaphora knowledge attention encodes. They defined a {\it confusion score}, which is the binary cross entropy of the normalized attention distribution between the anaphor and its candidate antecedents. They find that BERT attention does indeed encode some kind of coreference relationships that render the model to preferentially attend to the correct antecedent, though attention does not necessarily correlate with the linguistic effects we find in natural language.

Since the attention mechanism is nested in and jointly trained with the main model, 
interpretation methods based on attentions  thus 
belong to the {\it joint} line.

\subsection{Explanation Generation Based Interpretation}
\label{generation}
\begin{figure}[t]
  \centering
  \includegraphics[width=1\textwidth]{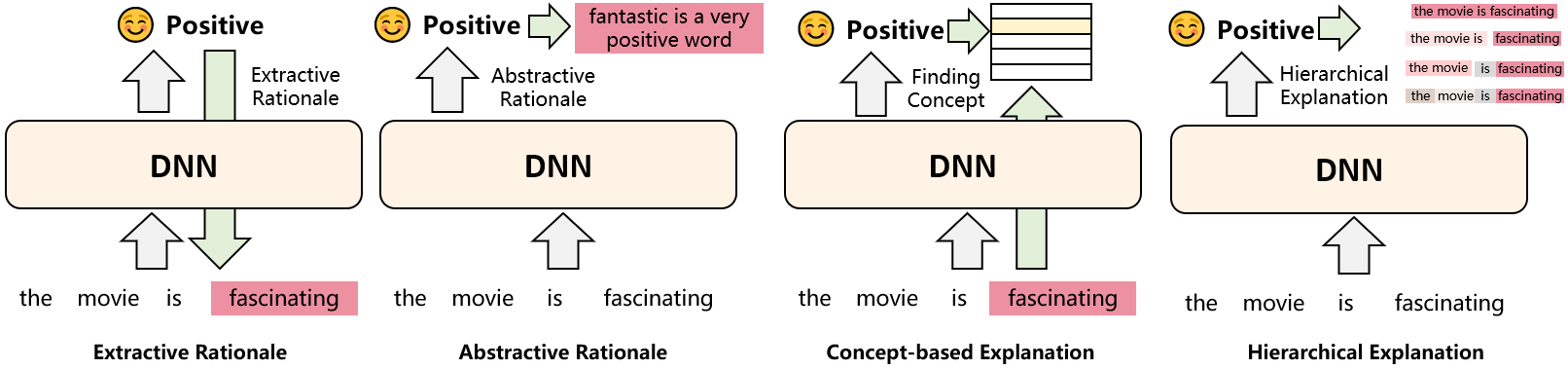}
  \caption{An overview of explanation generation methods. {\it First}: {\it extractive rationale generation} identifies rationales in the manner of extracting part(s) of the input. {\it Second}: {\it abstractive rationale generation} generates explanations in a sequence-to-sequence form by leveraging language models. {\it Third}: {\it Concept-based} methods interpret neural models by finding the most relevant concepts. {\it Forth}: {\it Hierarchy-based} methods produce semantic hierarchies of the input features.}
  \label{fig:test-generation}
\end{figure}

Saliency maps and attention distributions assign importance scores, which cannot be immediately transformed to human-interpretable explanations. This issue is referred to as
{\it explanation indeterminacy}. 
Explanation-based methods attempt to address this problem by generating explanations in the form of texts. 

Explanation-based methods fall into three major categories: {\it rationale-based} explanation generation, {\it concept-based} explanation generation and {\it hierarchical} explanation generation. Rationale-based explanation generation is 
further
sub-categorized into {\it extractive} rationale generation and {\it abstractive} rationale generation, which respectively refer to generating rationales within the input text in an extractive fashion, and generating rationales in a sequence-to-sequence model in a generative fashion. 
Concept-based methods aim at identifying the most relevant concepts in the concept space, where each concept groups a set of examples that share the same meaning as that concept: a concept represents a high-level common semantics over various instances. 
Hierarchy-based methods construct a hierarchy of input features and provide interpretation by examining the contribution of each identified phrase.
Explanation-based methods directly produce human-interpretable explanations in the form of texts, and thus are more straightforward to comprehend and easier to use.

Figure \ref{fig:test-generation} provides a high-level overview for the four categories of generation-based methods. 

\subsubsection{Rationale-based explanation generation}
A rationale is defined to be a short yet sufficient piece of text serving as justification for the model prediction \cite{lei2016rationalizing}. A {\it good} rationale should conceptually satisfy the following properties \cite{yu-etal-2019-rethinking}: 
\begin{itemize}
  \item {\it Sufficiency}: If the original complete input $x$ leads to the correct model prediction, then using only the rationale should also induce the same prediction.
  \item {\it Comprehensiveness}: The non-rationale counterpart should not contain sufficient information to predict the correct label.
  \item {\it Compactness}: The segments of the original input text that are included in the rationales should be sparse and consecutive, i.e., rationales should be as concise as possible.
\end{itemize}
We describe two styles of rationale generation, {\it extractive} and {\it abstractive}, and  review  representative works in the rest of this section.
\paragraph{Extractive rationale generation}~{}\\
\begin{figure}[t]
  \centering
  \includegraphics[width=1\textwidth]{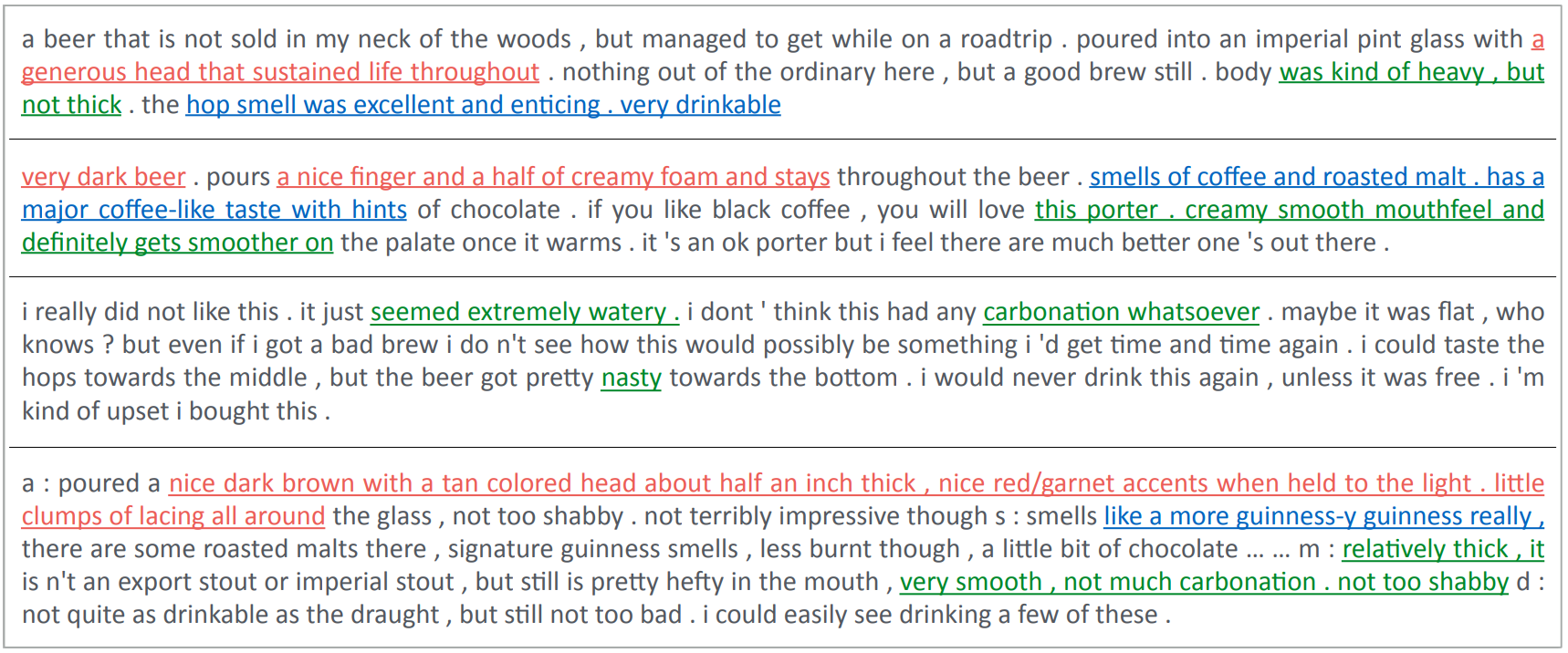}
  \caption{Examples of extracted rationales indicating the sentiments of various aspects (appearance:red, smell: blue, palate: green). This figure is from \cite{lei2016rationalizing}.}
  \label{fig:rationale-lei}
\end{figure}

\cite{lei2016rationalizing}  propose to extract rationales as interpretable justifications for the model prediction. The rationale generation process is incorporated as an integral part the overall learning problem, where the model consists of two modules: the {\it generator}, which specifies a distribution over possible rationales, and the {\it encoder}, which maps any text into target values. The generator and the encoder are jointly trained to optimize the objective that favors short, compact and sufficient rationales while preserving accurate predictions. More concretely, the encoder is used to digest the given input and output the predicted label. The encoder can be trained using a standard loss function, say, the squared error:
\begin{equation}
  \mathcal{L}_\text{enc}(x,y)=\Vert\text{enc}(x)-y\Vert^2_2
  \label{eq:rationale-encoder}
\end{equation}
At the meantime, a generator is employed to extract a subset of the original input text $x$ serving as the rationale. \cite{lei2016rationalizing} used binary variables $\{z_1,\cdots,z_i,\cdots,z_l\}$ where $l$ is the length of the input $x$ to denote whether each word $x_i$ should be selected or not. This process can be formalized as:
\begin{equation}
  z\sim\text{gen}(x)\triangleq p(z|x)=\prod_{i=1}^l p(z_i|x,z_{<i})
  \label{eq:rationale-generator}
\end{equation}
Equation \ref{eq:rationale-generator} indicates that the binary variable $z_i$ at time step $i$ is determined by the input $x$ as well as the previously generated binary variables $z_{<i}$, making the generation process sequentially dependent. With the extracted rationales at hand, we can rewrite Equation \ref{eq:rationale-encoder} into the following form:
\begin{equation}
  \mathcal{L}_\text{enc}(z,x,y)=\Vert\text{enc}(z,x)-y\Vert^2_2
  \label{eq:rationale-enc-gen}
\end{equation}
where $\text{enc}(z,x)$ means that the encoder is only applied to the rationales rather than the entire input text. Equation \ref{eq:rationale-enc-gen} encourages sufficiency of the extracted rationales.
Because we would like the extracted rationales to be as concise and compact as possible, we enforce the generator minimize the following regularizer:
\begin{equation}
  \mathcal{L}_\text{gen}(z)=\lambda_1\Vert z\Vert+\lambda_2\sum_i|z_i-z_{i-1}|
  \label{eq:rationale-gen}
\end{equation}
The first term encourages conciseness and the second term encourages consecutiveness. Overall, Equation \ref{eq:rationale-gen} compactness of the extracted rationales.

Combining Equation \ref{eq:rationale-enc-gen} and Equation \ref{eq:rationale-gen} together leads to the final training objective:
\begin{equation}
  \min_{\theta_\text{enc},\theta_\text{gen}}\sum_{(x,y)}\mathbb{E}_{z\sim\text{gen}(x)}[\mathcal{L}_\text{enc}(z,x,y)+\mathcal{L}_\text{gen}(z)]
  \label{eq:rationale-joint}
\end{equation}
Equation \ref{eq:rationale-joint} is optimized by doubly stochastic gradient descent, a variant of the REINFORCE algorithm \cite{DBLP:journals/ml/Williams92} as a sampled approximation to the gradient of Equation \ref{eq:rationale-joint}.
Figure \ref{fig:rationale-lei} shows several examples of the extracted rationales associated with the sentiments of different aspects. The proposed method successfully extracts interpretable rationales that can sufficiently explain the model predictions, and the extraction text snippets are generally concise and compact.

\begin{table}[t]
  \centering
  \begin{tabular}{p{9cm}c}
      \toprule 
      {\bf Improvement} & {\bf Representative works}\\\midrule
      Decoupling rationale selection and label prediction. & \cite{jain-etal-2020-learning}\\
      Improving masking strategies. & \cite{bastings2019interpretable,de2020decisions}\\
      Extracting rationales according to or conditioning on different classes. & \cite{chang2019game,yu-etal-2019-rethinking}\\
      Incorporating the information bottleneck theory. & \cite{paranjape2020information}\\
      Extracting rationales through matching. & \cite{swanson2020rationalizing,Huang_Chen_Du_Yang_2021}\\
      \bottomrule
  \end{tabular}
  \caption{Improvements of extractive explanation generation over \cite{lei2016rationalizing}.}
  \label{tab:extractive}
\end{table}

A large strand of works are motivated by \cite{lei2016rationalizing} and propose to improve extractive rationale generation from various perspectives. We introduce some typical directions in the following contents. A summary of these improvements is provided in Table \ref{tab:extractive}.

{\it Decoupling rationale selection and label prediction} \cite{jain-etal-2020-learning}. A drawback of the vanilla extractive rationale generation approach is the difficulty of training the two components -- the encoder and the generator simultaneously using only instance-level supervision, as shown in Equation \ref{eq:rationale-joint}. To tackle this issue, \cite{jain-etal-2020-learning} proposed to decouple the rationale selection process and the label prediction process, separately training the two modules and respectively applying them for rationale extraction and label prediction. The proposed pipeline works as follows: (1) train a support model to assign importance scores (Section \ref{saliency} and Section \ref{attention}) to input features, and discretize them into binary labels similar to the latent binary labels $z$ in \cite{lei2016rationalizing}; (2) treating the original text as input and the binarized labels as gold output, train an extraction model to extract rationales; (3) treating the extracted rationales as input and the gold class label as output, train a classifier for label prediction. The training pipeline introduces independent modules for rationale extraction and label prediction, naturally bypassing heavy recourse to reinforcement learning and mitigating the training difficulty.

\begin{figure}[t]
    \centering
    \begin{minipage}[t]{0.48\textwidth}
      \centering
      \includegraphics[width=1\textwidth]{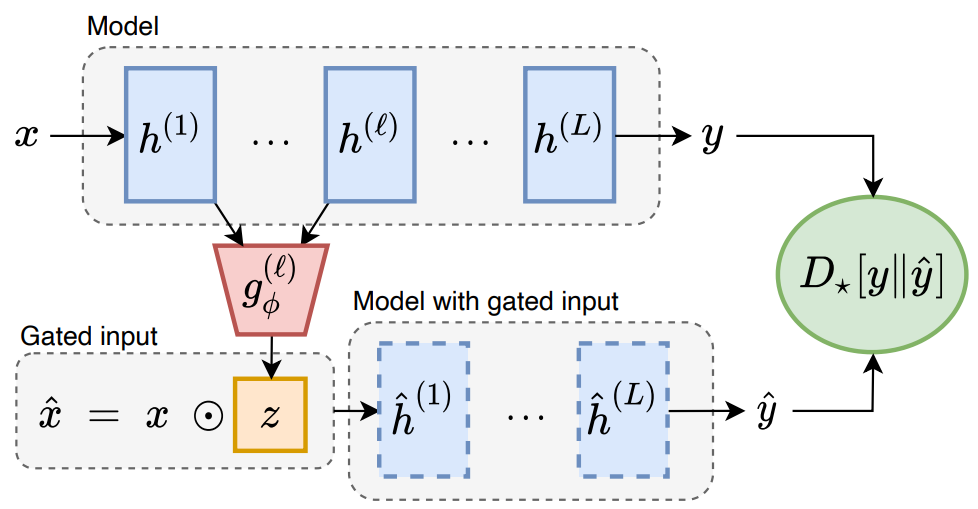}
    \end{minipage}%
    ~~~~~~~
    \begin{minipage}[t]{0.25\textwidth}
      \centering
      \includegraphics[width=1\textwidth]{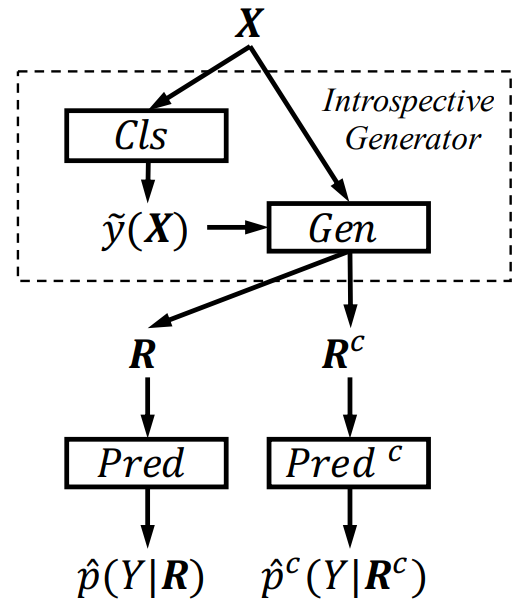}
    \end{minipage}
    \caption{Diagrams of the differentiable masking strategy (left, the figure is brought from \cite{de2020decisions}) and class-dependent method (right, the figure is brought from \cite{yu-etal-2019-rethinking}). {\it Left}: Hidden states up to layer $\ell$ are fed to a classifier to predict the mask, which is then used to run a forward pass with the input masked. The objective is to minimize the divergence. {\it Right}: An introspective generator first predicts the possible output $\tilde{y}$ and then generates rationales and rationale complements based on $x$ and $\tilde{y}$.}
    \label{fig:rationale-mask-class}
\end{figure}

{\it Improving masking strategies} \cite{bastings2019interpretable,de2020decisions}. This line of methods improve the masking strategy, i.e., how to assign binary variables as in \cite{lei2016rationalizing} to avoid sophisticated model training. The basic idea is to impose differentiability on the binary latent variables and the sparsity-inducing regularization, making solving Equation \ref{eq:rationale-joint} tractable even without REINFORCE. \cite{bastings2019interpretable} proposed to use the Kumaraswamy distribution \cite{kumaraswamy1980generalized}, a family of distributions that exhibit both discreteness and continuity, to allow for reparameterized gradient estimates and support for binary outcomes. \cite{de2020decisions} proposed to learn to mask out subsets while maintaining differentiability by pushing close the output predicted using the original input text and the output predicted using the extracted rationales. An overview of the proposed model in \cite{de2020decisions} is shown in Figure \ref{fig:rationale-mask-class} (left).

{\it Extracting rationales according to or conditioning on different classes} \cite{chang2019game,yu-etal-2019-rethinking}.
The common paradigm we have introduced above is to make an {\it overall} selection of a subset of the input that maximally explains the model decision, but rationales can be {\it multi-faceted} regarding different classes. To tackle this issue, \cite{chang2019game} proposed {\it class-wise rationales}, where different sets of rationales are generated to support the decisions on different classes.
\cite{yu-etal-2019-rethinking} similarly proposed to generate class-wise rationales, and more importantly they explicitly control the rationale complement via an adversarial model so as not to leave any useful information out of the rationale selection. The fine-grained control of rationale complement remedies the comprehensiveness property missed by the vanilla extractive model. An overview of the model proposed by \cite{yu-etal-2019-rethinking} is shown in Figure \ref{fig:rationale-mask-class} (right).

\begin{figure}[t]
  \centering
  \begin{minipage}[t]{0.5\textwidth}
    \centering
    \includegraphics[width=1\textwidth]{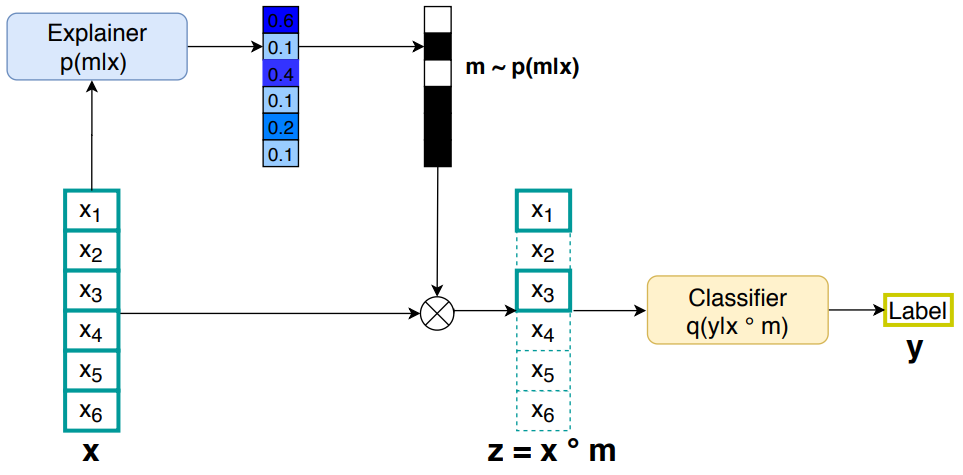}
  \end{minipage}%
  ~~~~~~
  \begin{minipage}[t]{0.25\textwidth}
    \centering
    \includegraphics[width=1\textwidth]{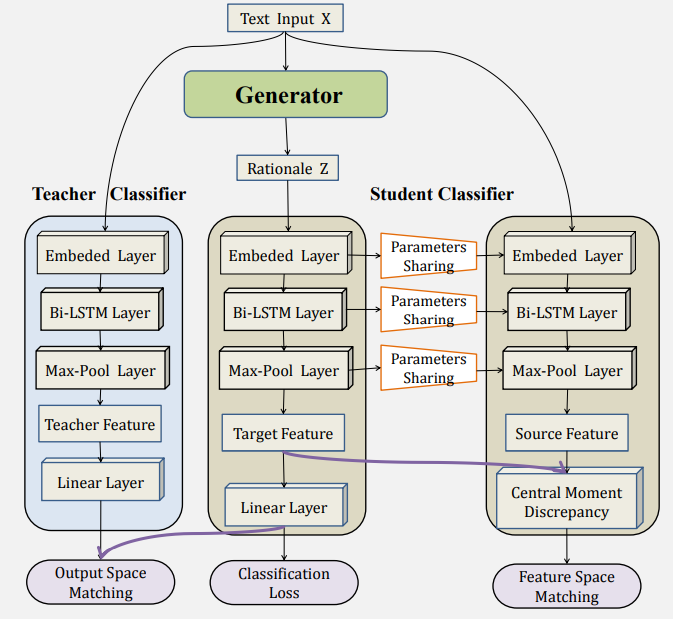}
  \end{minipage}
  \caption{Diagrams of the information bottleneck enhanced model (left, the figure is brought from \cite{paranjape2020information}) and the distribution matching model (right, the figure is brought from \cite{Huang_Chen_Du_Yang_2021}). {\it Left}: An explainer first extracts rationales from the input and then a classifier predicts the output based only on the rationale. The explainer leverages the Gumbel-Softmax reparameterization \cite{jang2016categorical} to reparameterize the Bernoulli variables. The full model is optimized by an information bottleneck guided objective. {\it Right}: The generator is enforced to generate rationales that match the full input feature in the feature space and match the full input prediction in the output space.}
  \label{fig:rationale-ib-matching}
\end{figure}

{\it Incorporating the information bottleneck theory} \cite{chen2018learning,paranjape2020information}. The reduction of the original input text inevitably results in performance degradation because the rationale complement usually contains non-essential but useful information. The trade-off between concise explanations (rationales) and high task accuracy can be better managed by incorporating the information bottleneck objective, where a rationale is expected to be (1) minimally informative about the original input and (2) maximally informative about the output class. A prior distribution provides flexible controls over the sparsity level, i.e., the fraction of the extracted rationales among the entire input text. Figure \ref{fig:rationale-ib-matching} (left) provides an overview of this method.

{\it Extracting rationales through matching} \cite{swanson2020rationalizing,Huang_Chen_Du_Yang_2021}. Intuitively, similar inputs should have similar rationales, and the extracted rationales should also get close to the original input in the feature space. Motivated by this intuition, better rationales can be extracted by matching similar inputs, or matching the distributions in the latent space. \cite{swanson2020rationalizing} proposed to employ Optimal Transport (OT) \cite{cuturi2013sinkhorn} to find a minimal cost alignment between a pair of inputs, providing a mathematical justification for rationale selection. \cite{Huang_Chen_Du_Yang_2021} did not perform matching in the textual level, but rather in the feature space and the output space. In the feature space, they imposed a regularizer that minimizes the central moment discrepancy \cite{zellinger2019central} between the full input feature and the rationale feature. In the output space, they minimized the cross entropy loss between the full input prediction and the rationale prediction, acting in a way similar to knowledge distillation \cite{hinton2015distilling}. Figure \ref{fig:rationale-ib-matching} (right) shows the method of \cite{Huang_Chen_Du_Yang_2021}.

\paragraph{Abstractive rationale generation}~{}\\
\begin{figure}[t]
  \centering
  \begin{minipage}[t]{0.4\textwidth}
    \centering
    \includegraphics[width=1\textwidth]{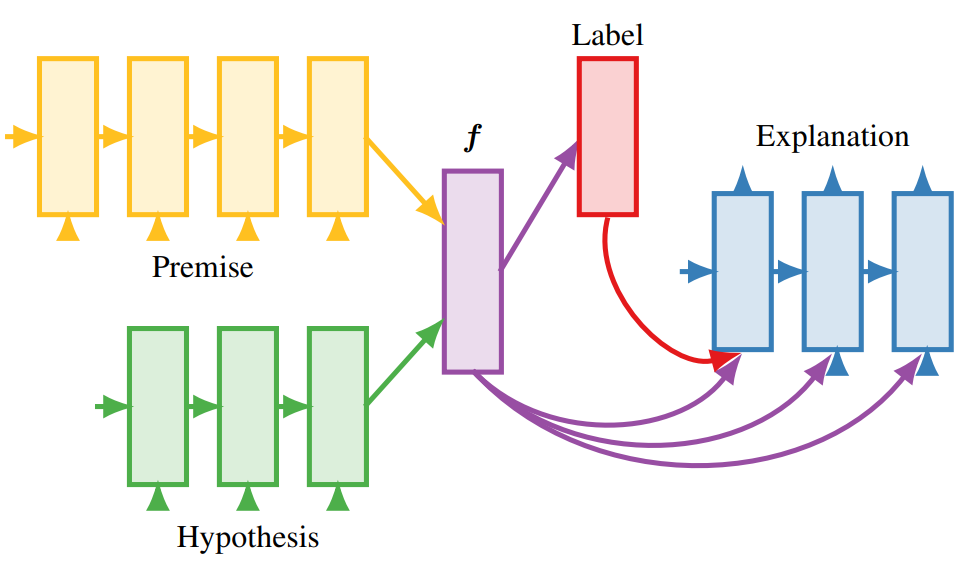}
  \end{minipage}%
  ~~~~~~~~
  \begin{minipage}[t]{0.38\textwidth}
    \centering
    \includegraphics[width=1\textwidth]{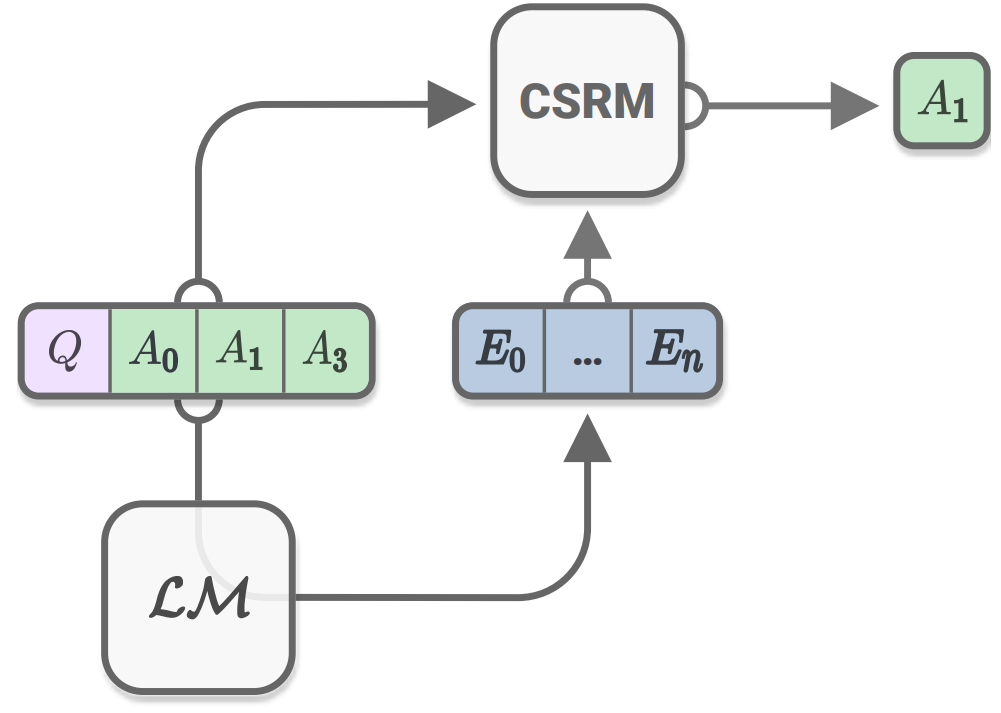}
  \end{minipage}
  \caption{Diagrams of the e-InferSent architecture (left, the figure is brought from \cite{camburu2018snli}) and the CSRM (Commonsense Reasoning Model) (right, the figure is brought from \cite{rajani2019explain}). {\it Left}: Given a pair of premise and hypothesis, the e-InferSent architecture jointly models the label and the corresponding explanation. {\it Right}: A language model is first used to generate explanations based on the question $Q$ and candidate answers $\{A_0,\cdots,A_n\}$, and CSRM is then used to predict the right answer.}
  \label{fig:rationale-snli-common}
\end{figure}

The extractive style rationale generation limits the space of rationales to the segments of the input text. In some cases, we would like the rationale to reflect the reasoning process of the model decision. For example, when predicting the input ``the movie is fantastic'' as label \texttt{positive}, the model should be able to reason that ``fantastic is a very positive word, and there is no negation that pushes the sentiment to the other side''. Extractive rationale generation would only, however, highlight the word ``fantastic'', but cannot explain why to select that word. In such cases, the model needs to provide {\it free-text} rationales by means of generating from scratch based on the input and the output. This type of rationale generation is called {\it abstractive} rationale generation, relative to {\it extractive} rationale generation.

The abstractive style rationale generation is usually implemented by natural language generation models, e.g., language models and sequence-to-sequence models. A common practice is to first craft datasets including the input instances as well as the corresponding human-annotated rationales, and then train language models in a supervised manner. For example, \cite{camburu2018snli} crafted the e-SNLI dataset, which is an extension of SNLI \cite{bowman2015large} focusing on providing human-labeled explanations for entailment relations; \cite{rajani2019explain} created CoS-E (Common Sense Explanations), an extension of CommonsenseQA \cite{talmor-etal-2019-commonsenseqa}, providing human explanations for commonsense reasoning. Both datasets offer direct supervision for model training, endowing the generation model with the ability to reason about its prediction and produce free-text rationales. Figure \ref{fig:rationale-snli-common} shows the model architectures proposed in \cite{camburu2018snli} and \cite{rajani2019explain}.

\begin{figure}[t]
  \centering
  \includegraphics[width=0.8\textwidth]{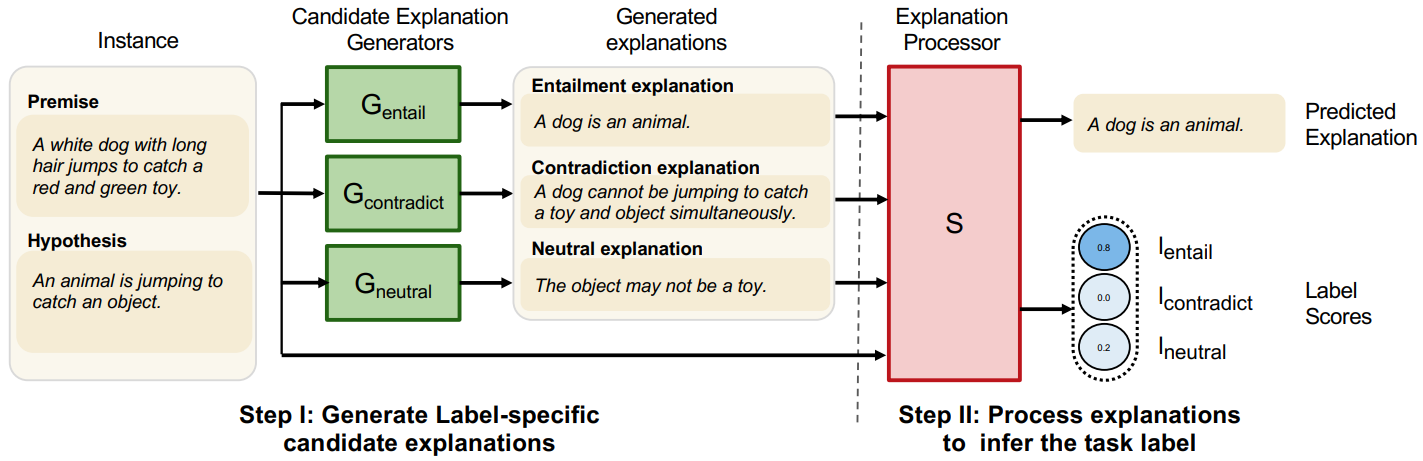}
  \caption{The workflow of the proposed model in \cite{kumar2020nile}. Given a pair of premise and hypothesis, the generator first produces label-specific explanations, which are then fed into the processor to infer label scores using the evidence present in these explanations. The figure is taken from the original literature.}
  \label{fig:rationale-nile}
\end{figure}

Similar to extractive rationale generation, the abstractive model can also generate label-specific explanations and then examine the faithfulness of the generated explanation with respect to the corresponding label \cite{liu2018towards,kumar2020nile,wiegreffe2020measuring}. A typical work comes from \cite{kumar2020nile}, the workflow of which is depicted in Figure \ref{fig:rationale-nile}. A merit that label-specific explanations offer is that it allows for re-examination on the generated rationales and on the whether these rationales can faithfully interpret the labels. This rationale generation approach supports testable explanations of the decisions and improves classification accuracy, even with limited amounts of labeled data.

An important advantage of abstractive rationale generation over extractive rationale generation is that abstractive models can be enhanced by large-scale language model pretraining \cite{raffel2019exploring,DBLP:journals/corr/abs-2004-14546}. Pretrained on massive unlabled general texts and finetuned on labeled rationale-guided data, the model is able to produce more coherent and human-identifiable explanations.

Extractive style and abstractive style rationale generation can be bridged via commonsense knowledge, rendering both types of explanations better \cite{majumder2021rationaleinspired}. The outlined pipeline works as follows: (1) extracting rationales most responsible for the prediction (using extractive rationale generation methods); (2) expanding the extracted rationales using commonsense resources; (3) using the expanded knowledge to generate free-text explanations. This pipeline naturally infuses commonsense knowledge into the process of rationale generation, therefore benefiting generating explanations more understandable to humans.

\subsubsection{Concept-based explanation generation}
\begin{figure}[t]
  \centering
  \includegraphics[width=1\textwidth]{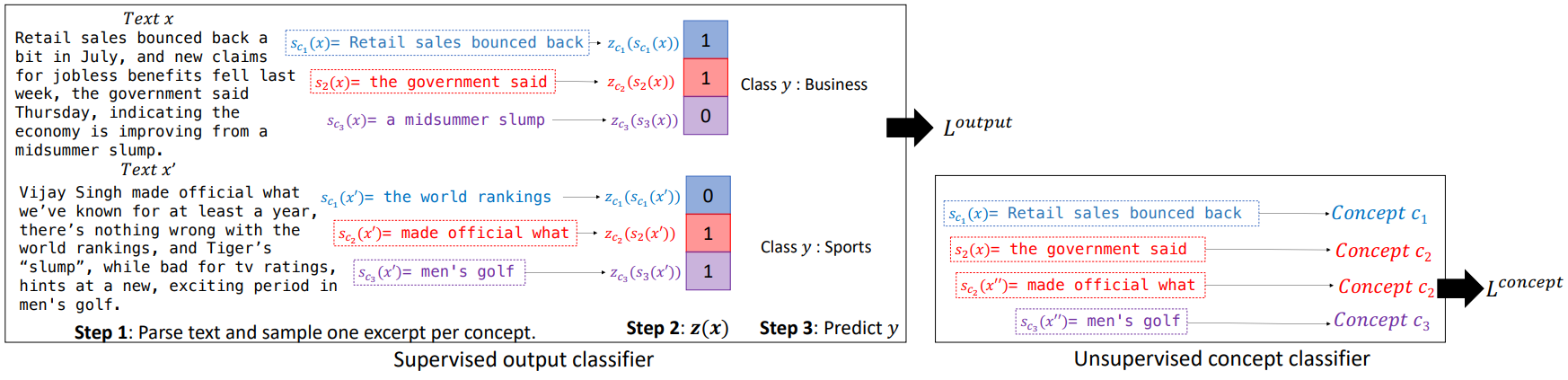}
  \caption{The workflow of the proposed EDUCE model in \cite{bouchacourt2019educe}. The model first samples one excerpt per concept, then determines whether the excerpt really activates the concept, and last feeds the binary representation for classification. An unsupervised concept classifier is used to enforce concepts consistency and to prevent overlap of concepts. The figure is taken from the original literature.}
  \label{fig:concept-educe}
\end{figure}

\begin{figure}[t]
  \centering
  \begin{minipage}[t]{0.35\textwidth}
    \centering
    \includegraphics[width=1\textwidth]{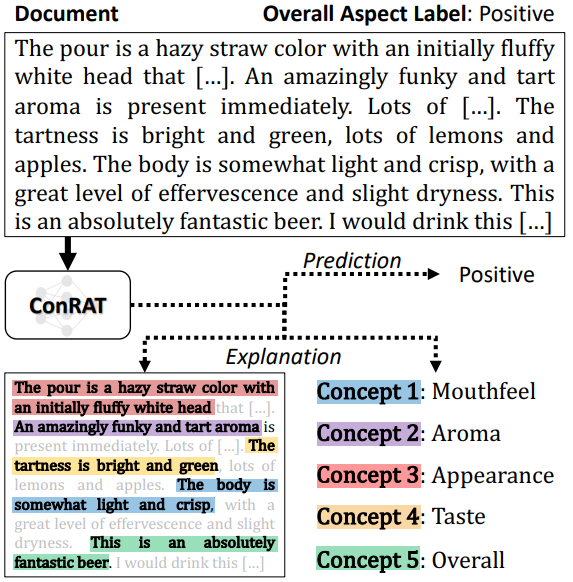}
  \end{minipage}%
  ~~~~~~~~
  \begin{minipage}[t]{0.38\textwidth}
    \centering
    \includegraphics[width=1\textwidth]{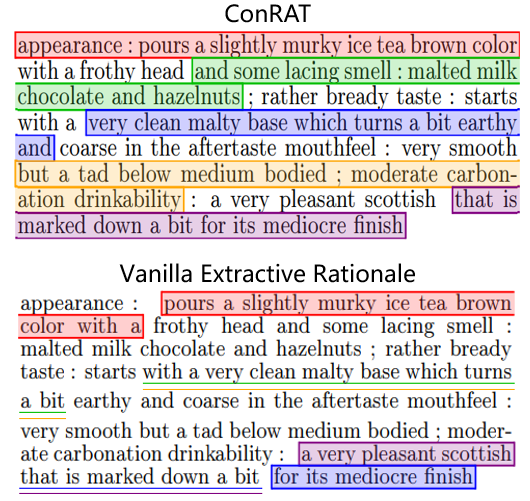}
  \end{minipage}
  \caption{{\it Left}: The pipeline of the proposed model ConRAT in \cite{antognini2021rationalization}. ConRAT first extracts a set of text snippets from the input and infers which ones are described as concepts (excerpts). Then the model explains the outcome with a linear aggregation of the concepts. {\it Right}: A comparison of the outputs produced by ConRAT and vanilla extractive rationale generation \cite{lei2016rationalizing}. The contents marked in red, green, yellow, blue and purple are respectively for appearance, aroma, palate, taste and overall sentiment. Vanilla extractive rationale produces spurious correlations while ConRAT extracts meaningful and non-overlapping concepts. The figures are taken from the original literature.}
  \label{fig:concept-conrat}
\end{figure}

Following the definition in \cite{bashier-etal-2020-rancc}, a {\it concept} \cite{alvarez2018towards,li2018deep,kim2018interpretability} is represented as a vector in the space which groups all the set of examples (rationales) that share the meaning of the concept. If the extracted rationale corresponds to a specific concept, that rationale is called an {\it excerpt} \cite{bouchacourt2019educe}.
For example, the phrases of ``fantastic movie'', ``enjoy the movie'' and ``the acting is very good'' share the same general concept \texttt{positive}, and all of these three phrases are excerpts that trigger the \texttt{positive} concept.
One  benefit provided by concept-based methods is the high-level clusterings of rationales that share the same general semantics. A concept can be simply the class label \cite{bashier-etal-2020-rancc}, e.g., the \texttt{negative} concept and the \texttt{positive} concept; a concept can also be more complex forms such as sub-categories of a class label \cite{bouchacourt2019educe}. For example, ask a user to classify the given text ``Retail sales bounced back a bit in July, and new claims for jobless benefits fell last week, the government said.'' The user could detect ``{\it retail sales}'' and, say, recognize it an \texttt{economy} concept; she would also note ``{\it the government said}'' and recognize it as an {\it politics} concept. Combining both concepts, the user can classify this input as label \texttt{Business}.  If the model correctly predicts the label \texttt{Business}, and at the same time extracts text pieces as well as their corresponding concepts, the model prediction would be more interpretable.
Extractive and abstractive rationale generations, however, cannot provide such high-level explanations, but only aim at interpret individual-specific inputs. 

Existing literature generates concepts in an unsupervised manner \cite{bouchacourt2019educe,bashier-etal-2020-rancc,rajagopal2021selfexplain,antognini2021rationalization}, where the concept is selected without any human annotation or priori. 
Two typical works come from \cite{bouchacourt2019educe} and \cite{antognini2021rationalization}. \cite{bouchacourt2019educe} proposed EDUCE (Explaining model Decisions through Unsupervised Concepts Extraction), which works in the following steps: (1) Compute $p(s|x,c)$, the probability of each excerpt $s$ in the input $x$ for each concept $c$, and sample one excerpt $s_c(x)$ according to the probability; (2) For each concept, determine whether the extracted excerpt is actually the trigger, i.e. computing $p(z_c|s_c(x),c)$. $z_c(s_c(x))=1$ means concept $c$ is detected as present, and absent otherwise; (3) Predict the class label $y$ using only the binary concept vector $z(x)=(z_1(s_1(x)),\cdots,z_C(s_C(x)))$, where $C$ is the number of concepts. In order for semantic consistency and to avoid overlap of the concepts, \cite{bouchacourt2019educe} jointly trained a concept classifier to categorize each excerpt into one of the $C$ classes, which exactly performs like a multi-label classifier. An overview is provided in Figure \ref{fig:concept-educe}.

\cite{antognini2021rationalization} proposed ConRAT, short for Concept-based Rationalizer, which could achieve stronger performance over \cite{bouchacourt2019educe} and infer non-overlapping and human-understandable concepts. The ConRAT model consists of two modules: a generator that finds the concepts, and a selector that detects whether a concept is present or absent. The key difference from \cite{bouchacourt2019educe} lies in the way of forming the concepts: \cite{bouchacourt2019educe} employed a simple concept classifier to decouple different concepts, whereas \cite{antognini2021rationalization} (1) favors the orthogonality of concepts and (2) minimizes the cosine similarity between concepts. Combining both regularizers, ConRAT is shown to produce accurate and interpretable concepts.
Figure \ref{fig:concept-conrat} (left) illustrates the pipeline of ConRAT. Figure \ref{fig:concept-conrat} (right) compares the rationales output by ConRAT and the vanilla extractive rationale generation method \cite{lei2016rationalizing}.

\subsubsection{Hierarchical explanation generation}
\begin{figure}[t]
  \centering
  \includegraphics[width=1\textwidth]{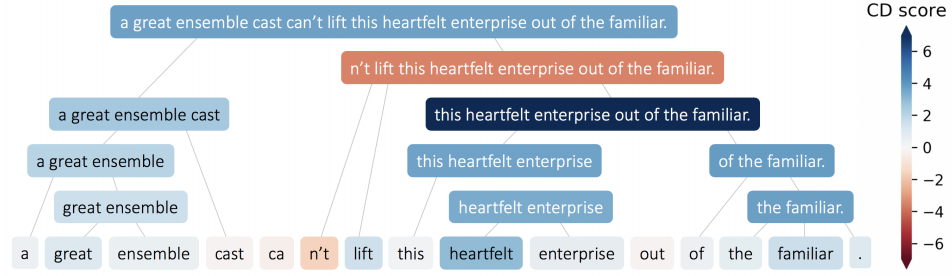}
  \caption{An example from \cite{singh2018hierarchical} illustrating how the explanation hierarchy works. Blue represents positive sentiment, white is neutral and red is negative. Higher rows display phrases associated with their sentiment scores identifies by the model.}
  \label{fig:hierarchy-acd}
\end{figure}

The third family of generation-based methods \cite{murdoch2018beyond,singh2018hierarchical,Jin2020Towards,chen2020generating} interpret model predictions by decomposing the input into tree-like segments at different granularity and computing the contribution of each segment to the model prediction. This forms a contextual hierarchy from the bottom level of words to the high level of phrases, enabling intuitive observations on how different parts of the input combine with each other and how their semantics influence the model prediction. Compared to rationale-based and concept-based methods, hierarchical explanation generation offers the most fine-grained interpretation, as they quantitatively measure the contribution of each detected phrase to the model prediction and exhibit the interpretable interactions between phrases.

The pioneer works \cite{murdoch2018beyond,singh2018hierarchical} for hierarchical explanation generation proposed {\it Contextual Decomposition} (CD), which decomposes the input into a hierarchy of phrases and assigns relevance scores signifying the contribution of each phrases to the model prediction. An example is shown in Figure \ref{fig:hierarchy-acd}. Concretely, a neural network $f(x)$ can be formalized as a composition of functions, followed by a softmax operator to derive the final prediction:
\begin{equation}
  f(x)=\text{softmax}(g(x))=\text{softmax}(g_L(g_{L-1}(\cdots(g_1(x)))))
\end{equation}
where $g(x)$ is the logits produced by the neural network. The logits, however, can be decomposed into the addition of two terms, $\beta(x)$ and $\gamma(x)$, where $\beta(x)$ captures the importance of the each input feature $x_i$ and $\gamma(x)$ captures the residue:
\begin{equation}
  g_i(x)=\beta_i(x)+\gamma_i(x),\forall i=1,2,\cdots,L
\end{equation}
The remaining question is how to compute $\beta_i(x)$ and $\gamma_i(x)$ for different neural architectures. \cite{murdoch2018beyond} proposed to disambiguate the interactions between gates (i.e. the output gate $o_t$, the forget gate $f_t$ and the input gate $i_t$) in LSTMs \cite{hochreiter1997long} by linearizing the update function. \cite{singh2018hierarchical} extended this idea to general DNN structures including convolutional, max-pooling, ReLU and dropout layers, etc. Besides, \cite{murdoch2018beyond} introduced the idea of {\it hierarchical saliency}, where a cluster-level importance measure is used to agglomeratively group features the model learns as predictive. The model is thus able to generate interpretable hierarchies more compliant to the model prediction.

\cite{Jin2020Towards} pointed out two properties CD-based methods should satisfy: (1) {\it Non-additivity}: the importance of a phrase should not simply be the sum of the importance scores of all the component words; (2) {\it Context independence}: the importance of a phrase should be evaluated independent of its context. Unfortunately, neither the methods of \cite{murdoch2018beyond} and \cite{murdoch2018beyond} satisfy both properties. To address this limitation, \cite{Jin2020Towards} proposed a new importance measure that satisfies both the non-additivity and context independence properties, producing consistently better explanations.

\cite{chen2020generating} designed a top-down model-agnostic method of constructing hierarchical explanations via feature interaction detection. Starting from the full input sentence, the model recursively divides the segments into smaller parts, until each segment contains only one word. \cite{chen2020generating} defined a {\it partition score}, which represents the degree of interaction between two successive segments. A small partition score means that the two segments are weakly interacted, i.e., partitioning these two segments would not cause a drastic semantic shift regarding interpretation. To evaluate the contribution of each new segment, \cite{chen2020generating} further defined a {\it feature importance score}, which is implemented as the difference between the predicted probability of the ground-truth label and the maximum predicted probability among other labels, given only the segment itself as input. This top-down recursive procedure can be applied to any neural model architecture and is free from complex mathematical derivation.

Though some explanation generation based methods can use post-hoc approaches to produce interpretation, e.g., saliency scores can be computed to extract the rationale, most existing works jointly train the interpreting model and the main model. We thus categorize explanation generation based interpretation into the {\it joint} line.

\subsection{Miscellaneous}
It is noteworthy that there are many other important miscellaneous works we do not mention in the previous sections. 
For example, numerous works have proposed to improve upon vanilla gradient-based methods \cite{smilkov2017smoothgrad,srinivas2019full,goh2021understanding};
linguistic rules such as negation, morphological inflection can be extracted by neural models \cite{prollochs2016negation,prollochs-etal-2019-learning,ruzsics-etal-2021-interpretability}; 
probing tasks can used to explore linguistic properties of sentences \cite{adi2016fine,10.5555/3241691.3241713,conneau-etal-2018-cram,hewitt-manning-2019-structural,jawahar2019does,hewitt-liang-2019-designing,chen2021probing}; 
the hidden state dynamics in recurrent nets are analysed to illuminate the learned long-range dependencies \cite{hermans2013training,karpathy2015visualizing,greff2016lstm,strobelt2016visual,kadar2017representation};
\cite{shi2016does,shen2017neural,shen2018ordered,kim2019unsupervised,drozdov2019unsupervised,shen2019ordered} studied the ability of neural sequence models to induce lexical, grammatical and syntactic structures;
\cite{jiang-etal-2019-explore,jiang-bansal-2019-self,asai2019learning,nishida2019answering,sap2019atomic,bosselut2019comet,ren-etal-2020-towards,jacovi2021contrastive} modeled the reasoning process of the model to explain model behaviors;
\cite{rothe2016word,park2017rotated,bravzinskas2017embedding,csenel2018semantic,zhao2018learning,shin2018interpreting,subramanian2018spine,panigrahi-etal-2019-word2sense,kozlowski2019geometry,dufter2019analytical,schwarzenberg-etal-2019-neural,hurtado-bodell-etal-2019-interpretable,lauretig-2019-identification,bommasani-etal-2020-interpreting} proposed to interpret word embeddings so that each dimension corresponds to a fine-grained sense and the value of that dimension represents the relevance of the sense to the word;
\cite{abujabal2017quint} drew on external knowledge base to derive a complete reasoning sequence from the natural language utterance to the final answer; 
\cite{toneva2019interpreting}  interpreted neural systems by combining natural language and the human brain imaging recordings.
We kindly refer the readers to the original literature of interest for details.

\section{Hybrid Methods}
\label{hybrid}
\begin{table}[t]
  \centering
  \begin{tabular}{p{7.5cm}p{7.5cm}}
  \toprule
  {\bf TreeLSTM}  & {\bf BERT} \\\midrule
  \multicolumn{2}{l}{{\it Test example 1}:  i  {\color{red}{\bf loved}} it ! } \\
  (a) if you 're a fan of the series you 'll  {\color{ao} {\bf love}} it.  &  (a) i  {\color{ao} {\bf  loved}} this film . \\
  (b) old people will  {\color{ao} {\bf  love}} this movie. & (b) you 'll probably  {\color{ao} {\bf  love}} it .\\
  (c) ken russell would  {\color{ao} {\bf love}} this .&(c)  a movie i  {\color{ao} {\bf loved}} on first sight \\
  (d) i  {\color{ao} {\bf loved}} this film . & (d) old people will  {\color{ao} {\bf love}} this movie. \\
  (e) an ideal  {\color{ao} {\bf love}} story for those intolerant of the more common saccharine genre . & (e) i  {\color{ao} {\bf like}} this movie a lot ! \\\midrule
  \multicolumn{2}{l}{{\it Test example 2}: 
  it 's not life-affirming -- its vulgar and mean , but i  {\color{red}{\bf liked}} it .} \\
  (a) i {\color{ao} {\bf like}} it . & (a) as an introduction to the man 's theories and influence , derrida is all but useless ; as a portrait of the artist as an endlessly inquisitive old man , however , it 's {\color{ao} {\bf invaluable}}  \\
  (b) the more you think about the movie , the more you will probably {\color{ao} {\bf like }}it . &(b) i {\color{ao} {\bf like}} it . \\
  (c) one of the {\color{ao} {\bf best}} , most understated performances of  jack nicholson 's career &(c) i liked the movie , but i know i would have liked it more if it had just gone that one step further .\\
  (d) it 's not nearly as fresh or {\color{ao} {\bf enjoyable}} as its predecessor , but there are enough high points to keep this from being a complete waste of time . &(d) sillier , cuter , and shorter than the first  as best i remember , but still a very {\color{ao} {\bf good}} time at the cinema \\
  (e)  i {\color{ao} {\bf liked}} it just enough . & (e) too daft by half ... but supremely {\color{ao} {\bf good}} natured . \\\bottomrule
  \end{tabular}
  \caption{An exmaple drawn from \cite{meng2020pair}.
  The most salient token in the test example regarding the golden label $y$ is marked in {\color{red}{\bf red}}.
  Each test example is paired with top 5 training examples that are the most responsible for the salient region by $ I(z,w_y(x_i))$ in Equation \ref{eq:hybrid-training}. For each extracted training example, its constituent token that is the most responsible for the salient region by $I(z_t, w_y(x_i))$ in Equation \ref{eq:hybrid-remove} is marked in {\color{ao} {\bf green}}.}
  \label{tab:hybrid}
\end{table}

Hybrid interpreting methods blend training-based interpretation and test-based interpretation, providing model interpretations from a global perspective.
A typical work comes from \cite{meng2020pair}, which jointly examines training history and test stimuli by associating influence functions and model gradients.

The first question hybrid methods answer is {\it which part(s) of the input example contribute most to the model prediction}. This question can be simply resolved by the techniques introduced in Section \ref{test}. However, as we would further like to interpret the detected part(s) from the training side, it is preferable to employ a differentiable method with which we can directly compute the influence functions regarding each training instance and the saliency part. \cite{meng2020pair} uses the {\it Vanilla Gradient} form for simplicity, and other gradient-based methods such as {\it Integrated Gradient} are also applicable. The saliency score for each input token $x_i$ can be expressed as follows:
\begin{equation}
  \text{Saliency}(x_i)=\Vert w_y(x_i)\Vert,~~~w_y(x_i)=\frac{\partial S_y(x)}{\partial x_i}
  \label{eq:hybrid-test}
\end{equation}
where $x_i$ is the corresponding word embedding of that word.
Recall in Equation \ref{eq:influence-chain} that the influence function $I(z,x)$ measures the influence of a training point $z$ on an input instance $x$, while in this case, we would like to associate a training point $z$ with a part of the input, say $x_i$. To this end, we can substitute $x$ in Equation \ref{eq:influence-chain} with the partial derivative $w_y(x_i)$ we just computed:
\begin{equation}
  I(z,w_y(x_i))=-\nabla_\theta w_y(x_i)^\top H_\theta^{-1}\nabla_\theta L(z;\hat{\theta})
  \label{eq:hybrid-training}
\end{equation}
Now for a given salient token $x_i$, Equation \ref{eq:hybrid-training} quantitatively measures the contribution of each training point $z$ to the detected part of the test input, given by $I(z,w_y(x_i))$. Moreover, the influence of perturbing a training example $z$ into $\tilde{z}$ on the salient part of a test example $x$ can also be measured:
\begin{equation}
  I(z,\tilde{z},w_y(x_i))\triangleq I(\tilde{z},w_y(x_i))-I(z,w_y(x_i))=-\nabla_\theta w_y(x_i)^\top H_\theta^{-1}(\nabla_\theta L(\tilde{z};\hat{\theta})-\nabla_\theta L(z;\hat{\theta}))
  \label{eq:hybrid-perturb}
\end{equation}
where $\tilde{z}$ is a perturbed version of the original training point $z$. When we implement perturbation as simple token removal, we can immediately quantify the influence of a certain token $z_t$ of a training point $z$ on the salience part $w_y(x_i)$:
\begin{equation}
  I(z_t,w_y(x_i))= I(\tilde{z},w_y(x_i))-I(z\backslash\{z_t\},w_y(x_i))
  \label{eq:hybrid-remove}
\end{equation}
Jointly examining training history and test stimuli offers a more comprehensive perspective for interpreting model decisions. Table \ref{tab:hybrid} shows an example drawn from \cite{meng2020pair} illustrating the identified salient part of the test input, the identified training points and the corresponding salient regions of these training points. Both models of TreeLSTM \cite{tai2015improved} and BERT \cite{devlin2018bert} successfully detect the salient part (i.e. ``loved'' and ``liked'') as well as the influential training points. \cite{meng2020pair} further applied the hybrid method to adversarial generation and model prediction fixing, exhibiting a wide utility of hybrid methods.

Hybrid methods combine influence functions and saliency scores and can be categorized into the {\it post-hoc} line.

\section{Open Problems and Future Directions}
\label{direction}
In spite of the variety of interpreting methods proposed for neural NLP, quite a few open problems still remain:
\begin{itemize}
  \item What interpretability exactly means and how to define interpretability: 
  existing works aim at proposing new and more effective interpreting models, but few clarifies what exactly interpretability is, 
  what the goal of interpretability is and how what philosophy we can take in general to reach interpretability \cite{miller2019explanation}. One can ``easily'' state that interpretable models are those that produce human-understandable results with reasoning behind predictions explicitly captured by humans. But quite a few questions arise: how we can formally define ``human-understandable''? What if the produced logic can not be recognized by some people but can be comprehended by others? 
  The lack of a clear   psychological definition and 
  a
  formal 
  mathematical formulation
   harms the development of further researches in neural NLP interpretation.
     \item How to evaluate interpreting methods: 
     The lack of formal definition for interpretability  leads to a lack of a widely accepted criterion for evaluation.
          Interpretability methods  are currently  evaluated by humans \cite{hase2020leakageadjusted,jacovi2020towards,wiegreffe2020measuring}, which are inevitably subjective:        
           For example, to evaluate saliency-based methods, 
           we can visualize the saliency maps generated by different interpreting methods to see whether they correctly identify the responsible part of the input for the model prediction; we can plot the attention distributions to interpret model decisions; we can also let the model to generate explanations. However, there is no such standardized metrics used to automatically evaluate model interpretability. Some datasets such as ERASER \cite{deyoung-etal-2020-eraser} have defined automatically metrics, but they still require human-labeled data for evaluation. 
           A robust and convenient criterion for the evaluation is urgently needed  to compare the effectiveness of different  interpretability models. 
    \item Whose interpretability are we talking about? Different populations or users' goals might affect which types of interpretability to use. In certain cases, end users or model practitioners may prefer different aspects of  interpretability that may not need to be related to the task itself. We urge future work on  interpretability to take into account more user centered design when developing interpretability methods and evaluations.  
    \item Go beyond classification tasks. Existing interpretability methods are mainly designed for text classification methods. How can we extend the aforementioned interpretability methods to other tasks such as named entity recognition, text generation or summarization? Certain interpretability methods such as attention-based interpretation might still be applicable to some extent, however, saliency-based ones may be fragile as it is hard to connect a specific token with the entire text sequence. 
  \item The inconsistency between different interpreting methods: It is common that existing interpreting methods contradict with each other when giving explanations, as shown in Figure \ref{fig:test-saliency}. The inconsistency of different interpreting methods makes users bewildered in terms of interpreting a model's decisions. 
   Future directions should explore the relations between different perspectives that different interpreting methods take, the factors that  make them  diverge, and the way of unifying them.
  \item The cost of model performance in exchange for interpretability: Neural interpretation aims at interpreting neural models while preserving model performance, complexity or resource consumption \cite{sun2020selfexplaining}. Some of existing methods improve model interpretability, but jeopardize model performance \cite{bertsimas2019price}. Studying such trade-offs between interpretability and model characteristics such as model performance is crucial for understanding the relationship between interpretability and model behaviors.
  \item The utility of interpretability:  The most important goal of developing interpretability methods is to use them, improving a neural model's 
    trustworthiness and controllability, helping improve and debug neural models, and facilitating the human AI collaboration. Currently, there hasn't been a systematic paradigm under which an interpretability model can be used in practice. 
    Future directions should explore the paradigm of how an interpretability model can be paired with its original model, 
     where AI systems provide not only decisions but also the human-interpretable reasoning process, helping users to judge how much they can trust a result given by an AI system  \cite{nguyen2018comparing,feng2019can,lai2019human,zhang2020effect,bansal2021does,jacovi2021aligning}. 
   \end{itemize}

\section{Conclusion}
\label{conclusion}
This survey showcases recent advances of research in interpreting neural NLP models. We first introduce the high-level taxonomy of existing interpreting methods. We have categorized related literature 
from two perspectives: 
training-based v.s. test-based, and
joint v.s. post-hoc. 
Then, we introduce typical works of each category in detail and shed light on their difference on the basic tools they use to interpret models. Last, we raise some open problems and future directions for this area in the hope of providing insights encouraging the community for a deeper understanding toward neural NLP interpretation.

\bibliography{interpret}
\bibliographystyle{acm}
\end{document}